# A Residual Guided strategy with Generative Adversarial Networks in training Physics-Informed Transformer Networks

Ziyang Zhang[1], Feifan Zhang*[1], Weidong Tang[2], Lei Shi[3], Tailai Chen[2]

[1]College of Science, China Agricultural University

[2]College of Information and Electrical Engineering, China Agricultural University

[3]College of Water Resources and Civil Engineering, China Agricultural University

*Correspondence: feifanzhang@cau.edu.cn

**Abstract**

Nonlinear partial differential equations (PDEs) are pivotal in modeling complex physical systems, yet traditional Physics-Informed Neural Networks (PINNs) often struggle with unresolved residuals in critical spatiotemporal regions and violations of temporal causality. To address these limitations, we propose a novel Residual Guided Training strategy for Physics-Informed Transformer via Generative Adversarial Networks (GAN). Our framework integrates a decoder-only Transformer to inherently capture temporal correlations through autoregressive processing, coupled with a residual-aware GAN that dynamically identifies and prioritizes high-residual regions. By introducing a causal penalty term and an adaptive sampling mechanism, the method enforces temporal causality while refining accuracy in problematic domains. Extensive numerical experiments on the Allen-Cahn, Klein-Gordon, and Navier-Stokes equations demonstrate significant improvements, achieving up to orders-of-magnitude reductions in relative MSE compared to baseline methods. This work bridges the gap between deep learning and physics-driven modeling, offering a robust solution for multiscale and time-dependent PDE systems.

## 1 Introduction

Nonlinear partial differential equations (PDEs) play a fundamental role in describing the dynamics of physical processes across various scientific and engineering fields, including fluid mechanics [1–2], solid mechanics [3–4], stochastic PDEs [5–6], and nonlinear optics [7–8]. In-depth investigation of these equations and their solution properties enable deeper understanding of physical phenomena and advances the development of mathematical physics [9]. However, despite significant progress in classical numerical methods—such as finite element methods, spectral methods [10], and meshless methods [11]—substantial challenges remain. Although traditional numerical solvers can achieve high accuracy, they often suffer from inevitable modeling errors, sensitivity to uncertainties, and high computational cost in practical applications [12–13].

In recent years, advances in deep learning have provided new perspectives for surrogate modeling of nonlinear systems, enabling data-driven solutions to forward and inverse PDE problems [14]. In 2017, Raissi et al. first proposed Physics-Informed Neural Networks (PINNs) [15], which incorporate physical laws directly into the loss function to approximate PDE solutions. This paradigm has since emerged as a promising alternative to conventional numerical solvers for a wide range of PDE problems [16]. Nevertheless, as a relatively new framework, PINNs exhibit inherent limitations that have motivated extensive subsequent research. Numerous extensions have been proposed, including novel network architectures such as Augmented PINNs [17], Conservative PINNs [18], Extended PINNs [19–22], Convolutional Variational PINNs [23], and Gradient-Enhanced PINNs [24]. Improvements have also been achieved through adaptive activation functions, such as the Self-Scalable Tanh [25], and adaptive hyperparameter strategies aimed at enhancing trainability [26–30]. In addition, software frameworks such as DeepXDE [31] and DeepONet [32] have facilitated broader adoption, while Fourier Neural Operators [33] have further advanced operator-learning-based approaches. Comprehensive benchmarking efforts, such as PINNacle [34], have also highlighted both the strengths and limitations of existing PINN variants. Despite these advances, a wide class of PDE problems remains challenging for PINNs, and their training failures are now recognized as systematic rather than pathological [35–36], underscoring the need to better understand the underlying causes and develop more robust training strategies. Importantly, these failures are increasingly understood as arising from the training dynamics themselves, rather than from insufficient model expressiveness or data availability.

One of the most critical and persistent failure modes of PINNs arises from the global averaging nature of the physics-informed loss. In particular, the residuals at key collocation points—often associated with sharp gradients, localized structures, or multi-scale features—can be severely diluted by the mean-squared-error formulation [37]. A key difficulty is that this is not simply a matter of adding more points or re-weighting the loss: without a mechanism that explicitly reallocates training effort, the optimization can keep spending most updates on “easy” regions that already satisfy the PDE approximately. As a result, even when the overall loss decreases steadily during training, important spatial or temporal features may remain under-resolved. This phenomenon reflects a fundamental mismatch between global loss minimization and the inherently localized difficulty of many PDE solution structures.

This limitation is closely tied to the sampling strategy used to evaluate the physics-informed loss. Conventional PINNs typically rely on static or uniformly random sampling of collocation points across the spatiotemporal domain. While computationally efficient, such strategies often fail to allocate sufficient sampling density to regions of high solution complexity, sharp transitions, or discontinuities. To address this issue, a growing body of work has focused on adaptive sampling techniques, which aim to refine collocation points based on suitable error indicators [24,38]. These approaches include stochastic sampling strategies such as Markov Chain Monte Carlo [39] as well as methods based on deep generative models [40]. In practice, such techniques usually require auxiliary probabilistic models or classical density estimators [41–42] to guide the sampling process.

Building upon these developments, adversarial learning has been increasingly explored as a mechanism to improve adaptive sampling. Representative examples include the adversarial adaptive sampling (AAS) framework [43], which seeks to jointly minimize the physics residual and optimize the training set. Compared with methods that directly rely on instantaneous residual values and are highly sensitive to noise, adversarial-learning-based approaches aim to learn a sampling distribution that concentrates probability mass in high-residual regions. This distributional perspective can alleviate repeated selection of identical collocation points and reduce the tendency toward local cycling and overfitting. Adversarial strategies for scientific computing are not entirely new. For example, Zang et al. [44] introduced weak adversarial networks that reformulated PDE solving as an operator norm minimization problem, while Zeng et al. [45] employed discriminator networks to evaluate the correctness of PINN predictions. The AAS framework further demonstrated the potential of adversarial learning for optimizing collocation point selection. However, despite these promising results, existing adversarial sampling methods still suffer from notable instability during training, often manifesting as high variance and sensitivity to noisy residual signals. From a broader perspective, the difficulty of residual-guided sampling is not merely algorithmic but structural: most existing strategies rely on pointwise residual ranking or threshold, which are intrinsically discontinuous operations. As a consequence, small perturbations in the residual—arising from numerical differentiation, stochastic optimization, or early-stage training instability—can lead to large and abrupt changes in the selected collocation set, undermining training stability.

Another common source of degradation in PINNs arises from violations of physical logic, particularly in time-dependent PDEs [46–47]. In such cases, global-in-time optimization may fail to respect the causal ordering of temporal evolution, leading to inaccurate representations of system dynamics. Importantly, inaccurate early-time predictions can contaminate residual evaluations at later times, thereby distorting residual-based sampling signals. As a result, residual misallocation and causality violations may reinforce each other, jointly impairing solution accuracy and the propagation of information from initial and boundary conditions into the domain of interest. Although existing studies have made progress in addressing adaptive sampling and temporal training strategies individually [48–50], these two issues are rarely considered in a unified manner, despite the fact that they typically do not occur independently during training. In practice, unstable residual-based sampling becomes more pronounced when residual signals themselves are corrupted by non-causal or premature optimization, further amplifying training instability.

To address this coupled failure mechanism, we propose a residual-guided training strategy for Physics-Informed Transformer Networks via Generative Adversarial Networks (GANs). The central objective of the proposed framework is to achieve stable and effective residual-guided sampling. To this end, residual-based sampling is reformulated as a distribution learning problem, in which a GAN is trained to generate collocation points that concentrate on physically meaningful and persistently difficult regions of the domain, rather than relying on pointwise residual ranking. Within this framework, the Physics-Informed Transformer is introduced primarily as a **c**ausality-aware residual provider, ensuring that residual signals used for sampling reflect genuine physical difficulty rather than spurious effects arising from incomplete convergence at earlier time steps. Through an alternating optimization strategy, GAN-generated collocation points are dynamically incorporated into the physics-informed loss, enabling smooth, stable, and adaptive refinement of the training set.

From this perspective, the proposed method advances residual-guided training by upgrading adaptive sampling from heuristic point selection to distribution learning and integrating it into a unified optimization framework. The main contributions of this paper are summarized as follows:

- We propose a residual-guided sampling framework based on GANs, which learns a smooth and stable sampling distribution over challenging spatiotemporal regions, avoiding the discontinuity and noise sensitivity inherent in residual ranking–based adaptive sampling methods.
- We integrate the proposed sampling strategy with a Physics-Informed Transformer architecture, which provides causality-consistent residual signals and supports stable residual evaluation in time-dependent PDEs.
- We demonstrate that the proposed framework yields improved stability, accuracy, and robustness across a range of PDEs with sharp interfaces, oscillatory solutions, and complex flow dynamics.

The remainder of this paper is organized as follows. Section 2 formulates the problem of solving PDE systems using deep neural networks and introduces the proposed

PhyTF-GAN framework. Section 3 presents comprehensive numerical experiments, including hyperparameter studies and comparisons with baseline methods. Finally, Section 4 concludes the paper and discusses future research directions.

## 2 Methodology

### 2.1 The Original PINNs

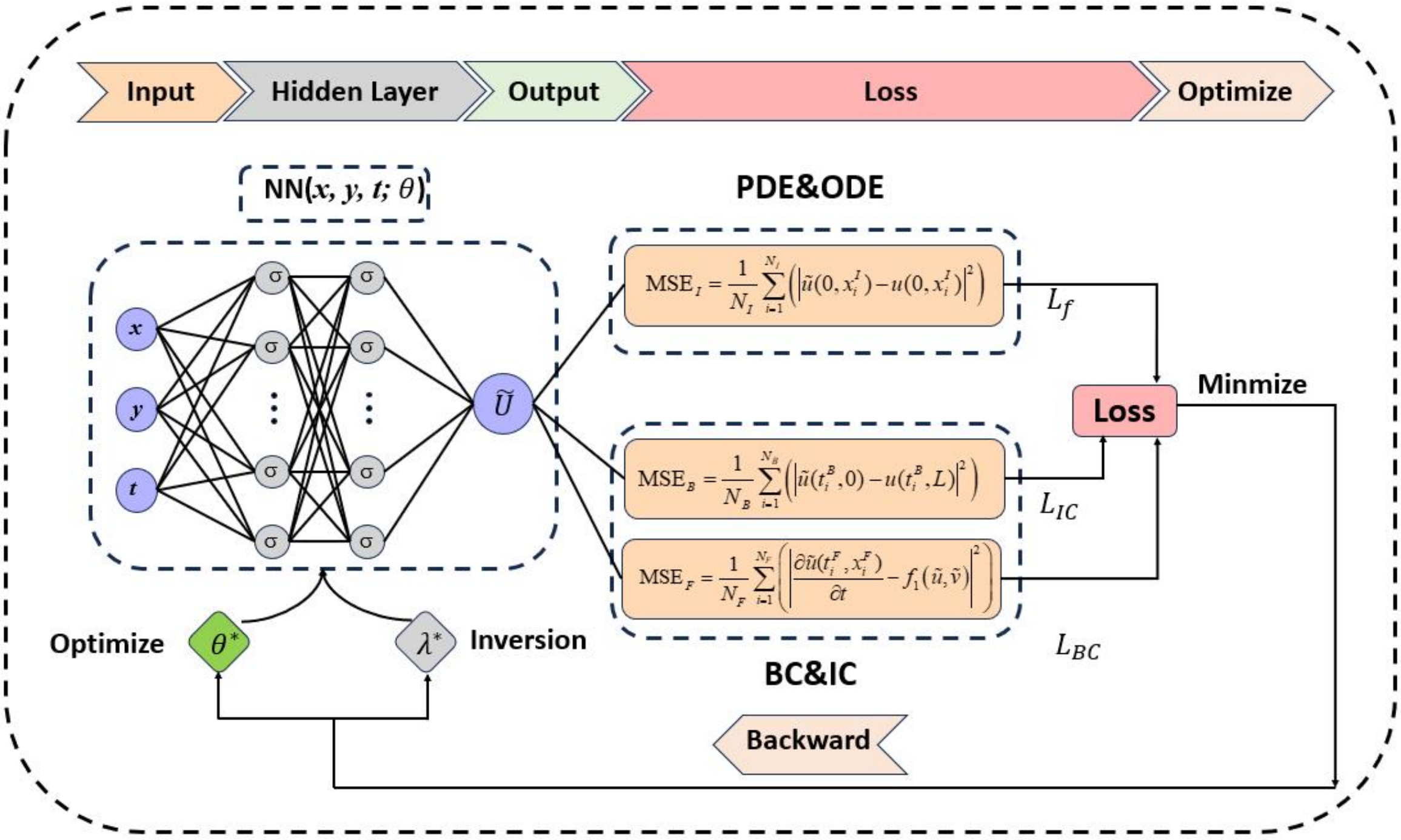

**Figure 1**. Structure of PINNs.

We initiate our discussion with a structured overview of PINNs. As depicted in **Figure 1**, the PINNs framework combines neural networks with physical governing equations to solve parameterized systems of PDEs. Herein, we focus on parameterized systems of PDEs in their general form. Let $u$ represent the dependent variables of the PDEs system, defined over spatial domain $\Omega$ and time domain $T$. A neural networks N ($u$ | $\theta$) with depth D (comprising an input layer, D-2 hidden layers, and an output layer) is designed as a parametric mapping from $R\hat{}m$ into $R\hat{}n$. To illustrate, consider a (1+1)-dimensional nonlinear PDE expressed as:

$$u_t + N[u] = 0, (x,t) \in \Omega \times T \tag{1.1}$$

$$u(x,t) = I(x,t), x \in \Omega \tag{1.2}$$

$$u(x,t) = \mathcal{B}(x,t), (x,t) \in \partial\Omega \times T \tag{1.3}$$

where $I$ and $\mathcal{B}$ stand for the initial condition and the boundary condition of equations, $u(x, t)$ denotes the solution, $x$ represents a space variable and $t$ represents a time variable. The loss function of PINNs consists of the partial differential equation loss (PDE loss), boundary condition loss (BC loss), and initial condition loss (IC loss), which are defined as:

$$\mathrm{MSE} = W_I \,\mathrm{MSE}_I + W_B \mathrm{MSE}_B + W_F \mathrm{MSE}_F . \tag{2.1}$$

Where

$$\mathrm{MSE}_I = \frac{1}{N_I} \sum_{N_I}^{i=1} | \tilde{u}(0, x_i^I) - u(0, x_i^I) |^2 , \tag{2.2}$$

$$\mathrm{MSE}_B = \frac{1}{N_B} \sum_{N_B}^{i=1} | \tilde{u}(t_i^B, 0) - u(t_i^B, L) |^2 , \tag{2.3}$$

$$\mathrm{MSE}_F = \frac{1}{N_F} \sum_{N_F}^{i=1} \left| \frac{\partial \tilde{u}(t_i^F, x_i^F)}{\partial t} - f_1(t_i^F, x_i^F) \right|^2 . \tag{2.4}$$

And $W_I$, $W_b$, $W_f$ are predefined hyper-parameters of weights.

The training procedure involves optimizing the neural network's parameter vector $\theta$ by minimizing a composite loss function that enforces governing physical laws and adherence to conditions at boundaries and initial states. Using automatic differentiation (AD), precise gradients of the network's predictions relative to input variables (e.g., space and time) are derived, which facilitates rigorous quantification of discrepancies in the governing PDE equations during optimization.

For time-dependent PDEs, physically meaningful solutions evolve sequentially in time, such that the state at a later time should be constrained by accurately learned states at earlier times. However, empirical studies by Wang et al. [53] have shown that vanilla PINNs trained with a global-in-time loss can exhibit non-causal training behavior, where residuals at later time instances are reduced even when the solution at earlier times remains inaccurate. From the perspective of the loss formulation in Eq. (2.4), all collocation points across the entire spatiotemporal domain contribute equally to the mean-squared residual. As a result, the optimization objective does not enforce any preference for resolving earlier time states before later ones. When non-causal training behavior occurs, residual reduction at later times may therefore take place without corresponding improvement at earlier times, which is inconsistent with the physical time-marching logic of the underlying PDE. To make the impact of this training behavior on the solution more explicit, for a given temporal discretization $\{t_i\}_{i=1}^{N_t}$ and spatial discretization $\{x_j\}_{j=1}^{N_x}$, the $\mathrm{MSE}_F$ loss (2.4) can be rewritten as:

$$\mathrm{MSE}_F = \frac{1}{N_t}\frac{1}{N_x}\sum_{N_t}^{i=1}\sum_{N_x}^{j=1}\cdot\left|\frac{\partial\tilde{u}(t_i^F, x_j^F)}{\partial t} - f_1(t_i^F, x_j^F)\right|^2, \tag{3.1}$$

Then, for any $\mathrm{MSE}_F$ could be approximated by discretizing $\frac{\partial\tilde{u}}{\partial t}$ through the forward Euler scheme [54]:

$$\mathrm{MSE}_F(t_i) \approx \frac{1}{N_x}\sum_{N_x}^{i=1}\left|\frac{\tilde{u}\left(t_i^F, x_j^F\right) - \tilde{u}\left(t_{i-1}^F, x_j^F\right)}{\Delta t} - f_1\left(t_i^F, x_j^F\right)\right|^2$$

$$\approx \frac{1}{\Delta t^2|\Omega|}\int_{\Omega}\left|\tilde{u}\left(t_i^F, x_j^F\right) - \tilde{u}\left(t_{i-1}^F, x_j^F\right) - \Delta t\cdot f_1\left(t_i^F, x_j^F\right)\right|^2 dx. \tag{3.2}$$

From the expression presented above, it could be seen that the minimization of $\mathrm{MSE}_F$ should be grounded in the accurate prediction of $\tilde{u}(t_i^F, \mathrm{x})$ and the preceding time $\tilde{u}\left(t_{i-1}^F, \mathrm{x}\right)$. As a result, inaccuracies at earlier time steps can propagate into the residual evaluation at later times. Even if the network optimizes later-time points first and obtains a more approximate estimate, such optimization does not lead to a truly accurate solution**.** In this case, large residuals at later times may partially reflect unresolved earlier states rather than purely local physical difficulty. Consequently, the residual signal becomes distorted, undermining its reliability as an indicator for adaptive sampling and potentially leading to systematic misallocation of collocation points.

**2.2 Residual-Awared PhyTF-GAN with Causality**

In this section, we propose a novel residual-aware and causality-preserving physics-informed framework, which integrates a Physics-Informed Transformer with a residual-oriented Generative Adversarial Network, forming a unified architecture termed PhyTF-GAN. The proposed PhyTF-GAN framework, as illustrated in **Figure 2,** is specifically designed to address the aforementioned challenges that arise in conventional PINNs, particularly the optimization difficulty in localized high-residual regions and the violation of temporal causality in time-dependent PDEs. By explicitly incorporating causality-aware mechanisms, PhyTF-GAN stabilizes residual evaluation by prioritizing earlier time steps during optimization. Meanwhile, an adaptive residual-guided sampling strategy based on adversarial learning is employed to progressively allocate training samples toward persistently under-resolved spatiotemporal regions, thereby improving convergence behavior across the entire domain.

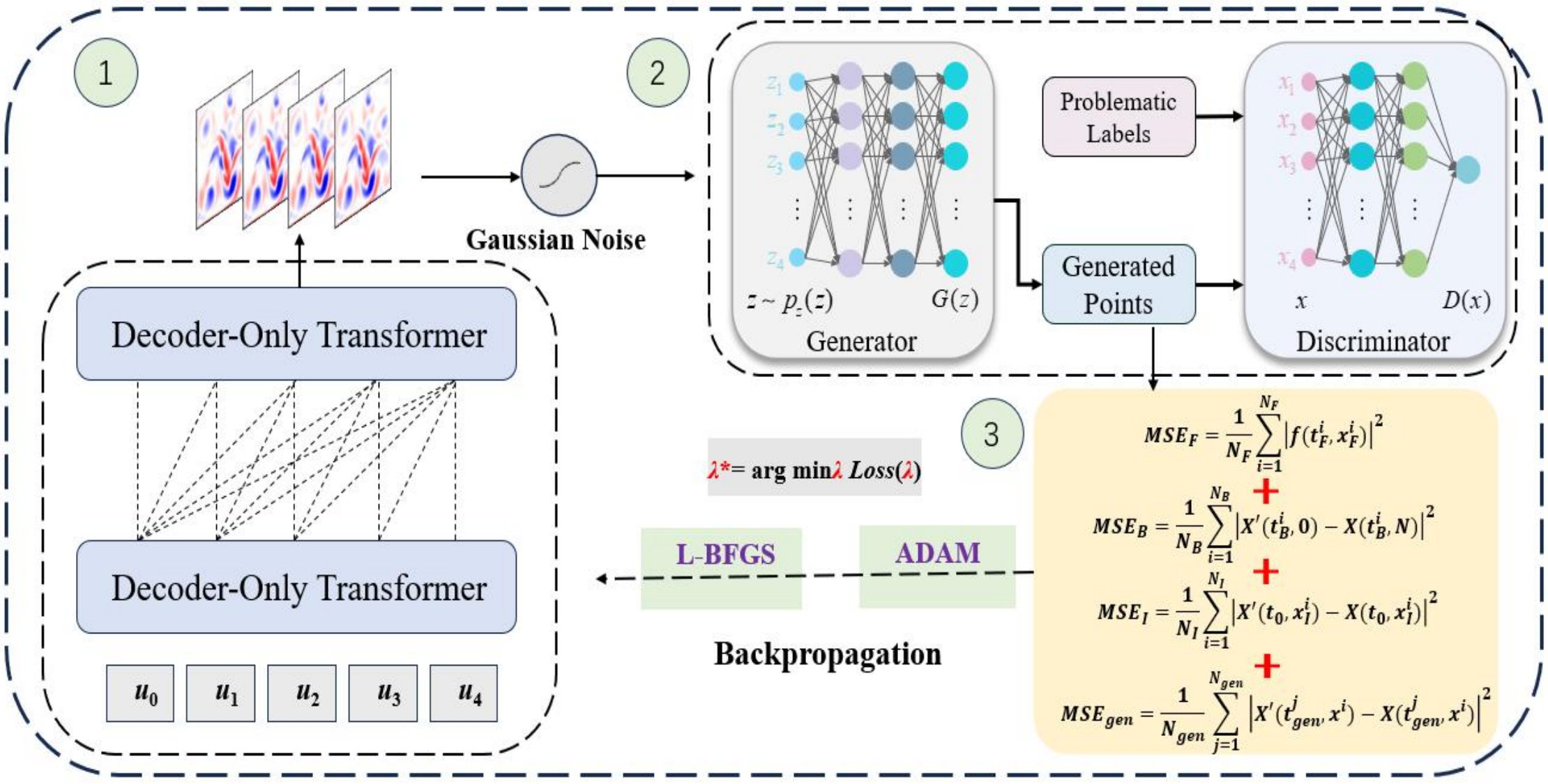

**Figure 2**. Structure of PhyTF-GAN.

To start with, we adopt the Physics-Informed Transformer as the baseline model. Owing to the built-in regression mechanism of the transformer, this baseline model inherently captures causal relationships over time. By integrating this model with a GAN and adopting an alternating training strategy, we aim to learn a sampling distribution that concentrates on persistently under-resolved regions. The GAN-generated samples are then incorporated into the physics-informed loss to guide the optimization of the transformer. As a result, the proposed framework provides a unified mechanism for integrating causality-aware residual evaluation with adaptive sampling.

### 2.2.1 Physics-Informed Transformer with Causality

In this part, we present the architecture of Physics-Informed transformer as shown in **Figure 3**. Here, we only take the initial value at time $t_0$ as input. After compressing it into a format suitable for the transformer, the network gradually predicts the values for the subsequent steps. More specifically, during one training iteration, the model first predicts the value at $t_1$ using the value at $t_0$ as input. Next, it combines the values of $t_0$ and $t_1$ to predict $t_2$, and this process continues in a time-marching manner. By sequentially leveraging all preceding time steps [$t_0$, $t_1$, ..., $t_{n-1}$], the model progressively solves for the values across time steps [1, n].

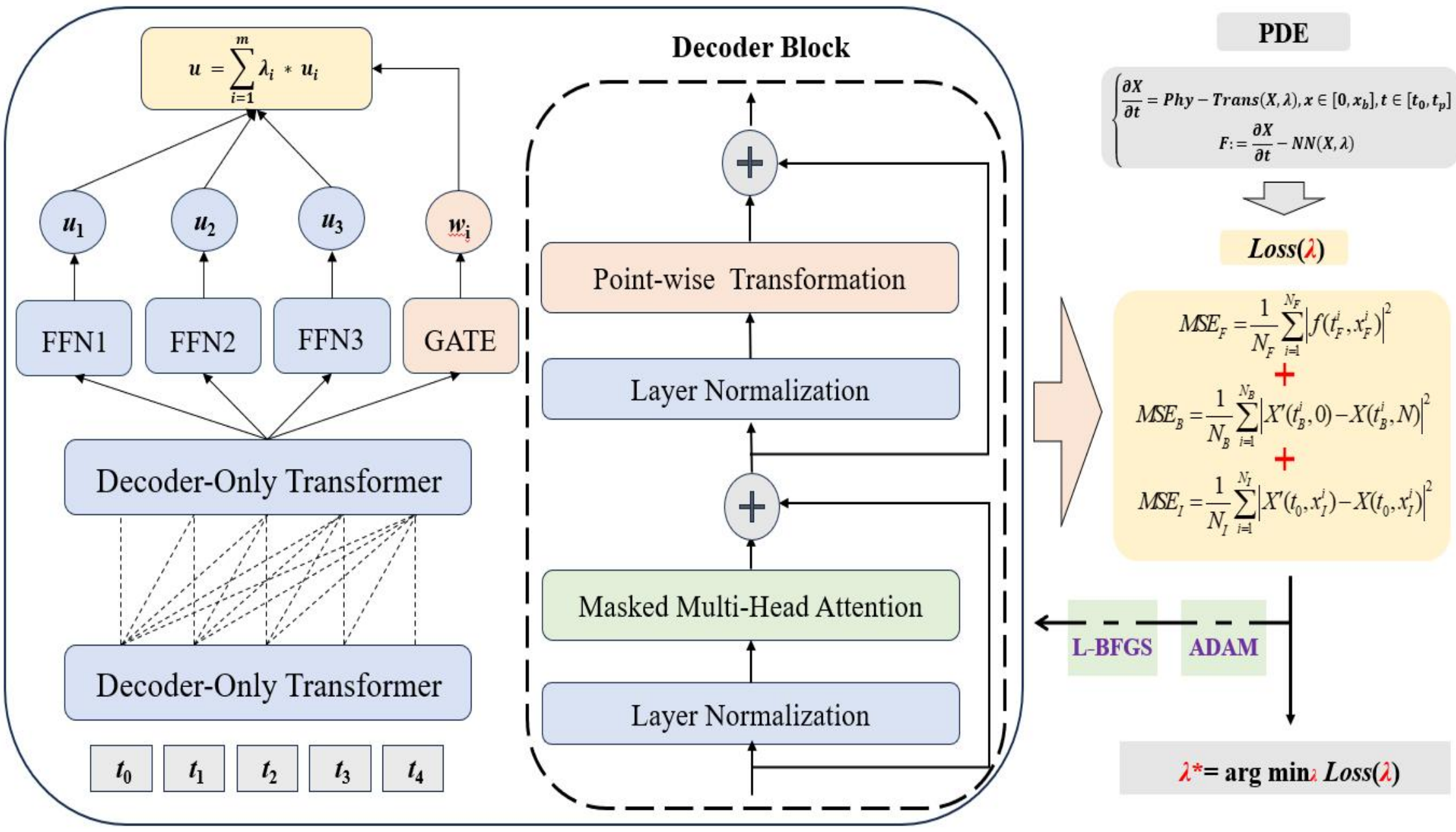

**Figure 3**. Structure of Phy-Transformer.

For time-dependent PDEs, it is a well-founded approach to introduce a time-marching method. By incorporating time series into neural networks, we can, to a certain degree, integrate the inherent causal relationship into the neural network framework. In this way, the neural network can better capture the temporal evolution characteristics of the PDEs, making the model more consistent with the temporal structure implied by time-dependent systems. However, the commonly seen attention-based and recurrent neural architectures, currently both have their own drawbacks in the application of solving time-dependent PDEs. Attention-based models require substantial data and disregard the underlying sequential causality inherent in the physical simulations, while Recurrent Neural Networks (RNNs) are constrained by the exploding and vanishing gradients problem (EVGP) [55].

Therefore, it is indeed necessary to adjust the existing framework for solving PDEs. Although the self-attention mechanism in the transformer encoder does not explicitly enforce temporal causality by default, the decoder-only transformer naturally supports autoregressive generation, in which predictions at each time step depend only on previously generated states. Actually, decoder-only transformer has been extensively used in natural language processing (NLP) tasks, with the most well-known example being the GPT framework [56-57].

The core idea of the decoder - only transformer is to gradually generate the output sequence through autoregression and a masked attention mechanism. In an autoregressive generation task, the model generates each token in the sequence step by step, only predicting the next token based on the already generated tokens and is prohibited from "peeking" at future tokens. This requirement is met by masking the future tokens with zero during the attention calculation, which effectively excludes the influence of future tokens and ensures the sequential and causally consistent nature of the generation process. However, such a mechanism can only ensure that the output is generated in accordance with the causal relationship. But during the optimization process, in fact, all the tokens in the entire batch are optimized together. That is to say, it is highly likely that the tokens at the later positions are optimized first, which is inconsistent with the sequential dependency discussed above**.** To address this, we introduce a causal penalty term to the loss function, explicitly guiding the optimization process to respect sequential dependencies.

Define a causal mask M∈{0,1}T, where $M_t = 1$ only if one condition holds: the loss at step *t* is below a threshold ϵ. This ensures no step can be marked as "solved" until all prior causal steps are stable. The penalty term $P_{\text{causal}}$ quantifies violations of this order: it counts instances where a later step *t'* is satisfied ($M_{t'} = 1$) while at least one earlier step $t < t'$ remains unsatisfied ($M_t = 0$). Mathematically:

$$P_{\text{causal}} = \sum_{t=1}^{T-1} \sum_{t'=t+1}^{T} \left(1 - M_t\right) M_{t'} \tag{4.1}$$

Then we put this penalty into the total loss:

$$L_{\text{total}} = L_{\text{pde}} + \lambda \cdot P_{\text{causal}}. \tag{4.2}$$

The number $\lambda$ controls how strong the causal rule is. By adjusting $\lambda$, we prioritize the optimization of early, foundational steps-crucial for establishing a reliable starting point—over later, dependent steps. This prevents the model from prematurely focusing on large losses in later steps and ignoring causal precedence.

### 2.2.2 Residual-Oriented PINNs with GAN Integration

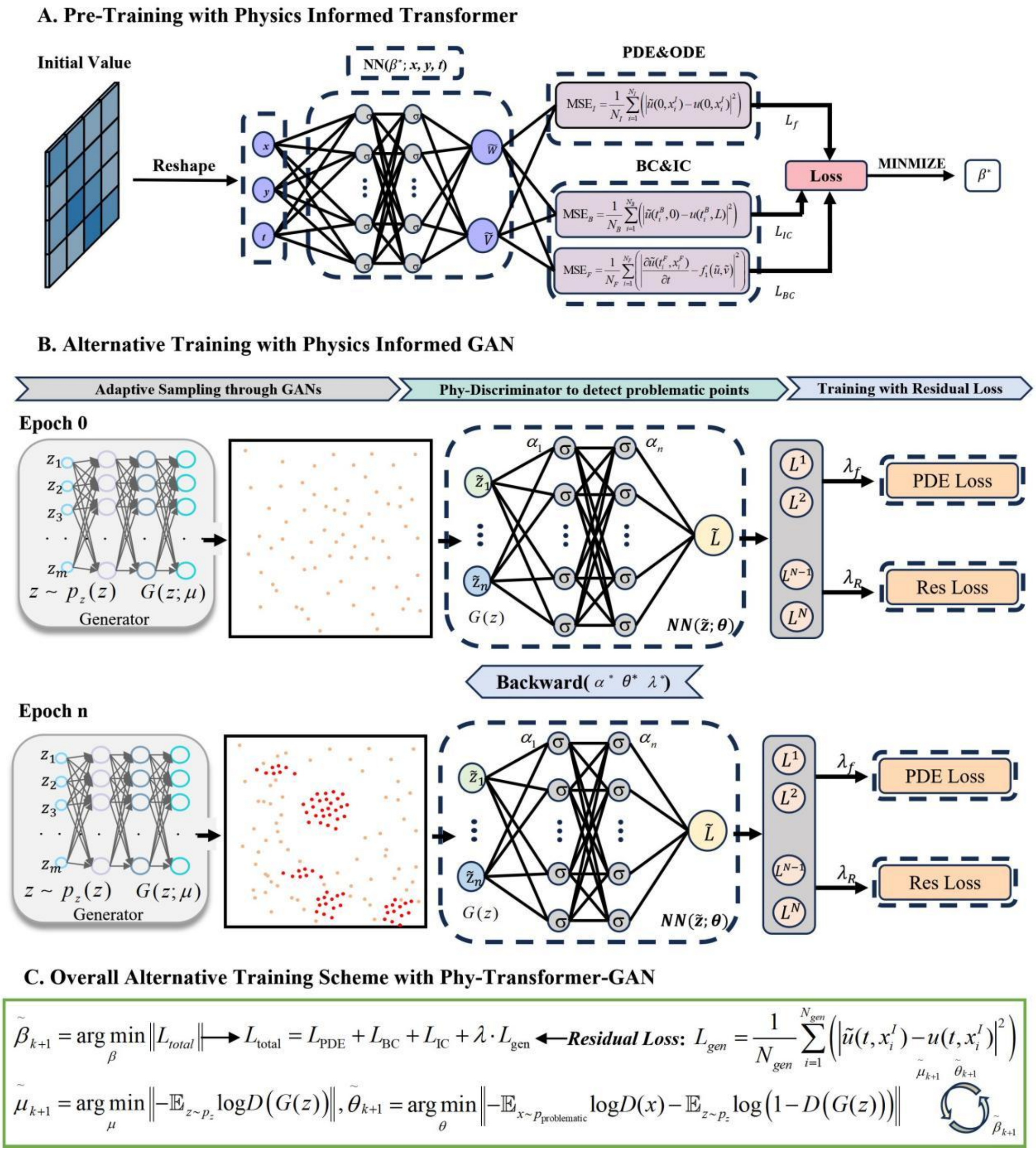

**Figure 4**. Training process of PhyTF-GAN.

In this stage, we focus on dealing with the problematic areas. To address this challenge, we propose a Residual-Oriented Generative Adversarial Networks (GAN) framework that dynamically identifies and prioritizes spatiotemporal points where the model struggles. This builds on the original GAN design by incorporating decision-making via physical laws, enabling the generator to learn adaptive sampling strategies guided by both discriminative feedback and physical loss.

Unlike a conventional GAN, which transforms Gaussian noise into samples of a target data distribution, our generator produces spatiotemporal points $(t, x, y) \in \Omega$, and its input latent vector is explicitly augmented by features derived from the PDE residuals. This construction, as shown in **Figure 4,** introduces physically informed decision-making into the generative process: the discriminator labels points as "problematic" or "normal" based on the PDE residual of the current Phy-Transformer, and the generator learns a sampling distribution that adaptively concentrates density on the difficult regions of the domain.

To establish the theoretical stability of this sampling mechanism, we first introduce the notation and recall several classical tools from real analysis. Let $Z \subset \mathrm{R}^m$ denote the latent space of the generator, $\Omega \subset \mathrm{R}^d$ denote the physical domain of the PDE, with typical points $z \in Z$ and $x=(t, x, y) \in \Omega$. The Euclidean norm is written as $// \cdot //$, with $// \cdot //$ and $// \cdot //$ denoting the $l^2$ and $l^\infty$ norms when necessary. The generator is written as a finite-depth feedforward map $G: Z \rightarrow \Omega$, and its Jacobian at $z \in Z$ is denoted by $J_G(z)$, satisfying the differential relation

$$dG = J_G(z)\, dz. \tag{5.1}$$

***Theorem 2.1***: *A continuous function defined on a compact set is bounded and attains its maximum and minimum.*

***Theorem 2.2***: *The composition of Lipschitz functions is Lipschitz, with constant equal to the product of the individual constants.*

Having recalled these foundational results, we proceed with the stability analysis of the proposed residual-oriented sampling scheme. The governing PDE is written as

$$\mathcal{N}[u](x) = f(x), \qquad x \in \Omega,. \tag{5.2}$$

and its PINN approximation is denoted $u_\theta$. The exact residual is

$$R_\theta(x) = \mathcal{N}\left[u_\theta\right](x) - f(x),. \tag{5.3}$$

and the ideal loss is

$$.\ell_{\text{clean}}(x;\theta)=\phi\left(R_{\theta}(x)\right), \tag{5.4}$$

where $\phi(s)=\frac{1}{2}s^2$ or $|s|$. In numerical computation, an approximate operator $\tilde{N}$ is used, producing the numerical loss

$$\ell_{\text{num}}(x;\theta)=\phi\left(\tilde{\mathcal{N}}\left[u_{\theta}\right](x)-f(x)\right).. \tag{5.5}$$

We define the loss perturbation

$$\eta(x;\theta)=\ell_{\text{num}}(x;\theta)-\ell_{\text{clean}}(x;\theta).. \tag{5.6}$$

By continuity of the above maps and compactness of $\Omega$ and the parameter set $\Theta$, Theorem 2.1 immediately yields the uniform bound

$$\left|\eta(x;\theta)\right|\leq\varepsilon,\qquad\forall(x,\theta)\in\Omega\times\Theta,. \tag{5.7}$$

and therefore, after discretization,

$$\ell_{\text{num}}=\ell_{\text{clean}}+\eta,\qquad\|\eta\|_{\infty}\leq\varepsilon. \tag{5.8}$$

The latent vector used by the generator is not the base noise $z$, but an augmented latent variable

$$z_{\text{aug}}=\Psi\left(z,\ell_{\text{num}}\right) \tag{5.9}$$

where $\Psi$ is a residual encoder. Since $\Psi$ is a finite-depth feedforward network with Lipschitz activation functions, it is globally Lipschitz on bounded sets, so there exists $L_{\Psi}>0$ such that

$$\|\Psi\left(z,\ell_{1}\right)-\Psi\left(z,\ell_{2}\right)\|\leq L_{\Psi}\|\ell_{1}-\ell_{2}\|. \tag{5.10}$$

Applying the generator's Lipschitz property gives

$$\|G\left(z_{1}\right)-G\left(z_{2}\right)\|\leq L\|z_{1}-z_{2}\|,\quad\forall z_{1},z_{2}\in Z. \tag{5.11}$$

Combining Eq. (5.8), (5.10) and (5.11) yields

$$\begin{aligned}\|G\left(\Psi\left(z,\ell_{\text{num}}\right)\right)-G\left(\Psi\left(z,\ell_{\text{clean}}\right)\right)\| &\leq L\|\Psi\left(z,\ell_{\text{num}}\right)-\Psi\left(z,\ell_{\text{clean}}\right)\|\\ &\leq LL_{\Psi}\|\ell_{\text{num}}-\ell_{\text{clean}}\|\\ &\leq LL_{\Psi}\varepsilon.\end{aligned} \tag{5.12}$$

To extend this pointwise stability to distribution-level stability, define

$$X_{\text{clean}}=G\left(\Psi\left(Z,\ell_{\text{clean}}\right)\right),\quad X_{\text{num}}=G\left(\Psi\left(Z,\ell_{\text{num}}\right)\right), \tag{5.13}$$

with $Z \sim p(z)$. To transfer pointwise stability to distribution-level stability, we apply the standard coupling characterization of the Wasserstein-1 distance together with Eq. (5.12), we obtain

$$W_1\left(\mu_{\text{clean}}, \mu_{\text{num}}\right) \leq L L_{\psi} \varepsilon . \tag{5.14}$$

Thus, the continuous sampling distribution is Lipschitz-stable under bounded noise in the PDE loss. We now analyze the discrete sampling probabilities. A scoring network $g$ outputs logits $a(\mathrm{l}) = g(\mathrm{l})$. As it is also a finite-depth feedforward network with Lipschitz activation functions, $g$ is Lipschitz with constant $L_g > 0$:

$$\| g\left(\ell_1\right) - g\left(\ell_2\right) \|_2 \leq L_g \| \ell_1 - \ell_2 \|_2 . \tag{5.15}$$

The temperature-scaled SoftMax

$$p_i(\ell) = \frac{\exp\left(\mathrm{a}_{\mathrm{i}}(\ell) / \tau\right)}{\sum_{\mathrm{j}=1}^{\mathrm{N_f}} \exp\left(\mathrm{a}_{\mathrm{j}}(\ell) / \tau\right)} \tag{5.16}$$

has Jacobian

$$\frac{\partial p_i}{\partial a_j} = \frac{1}{\tau} p_i\left(\delta_{ij} - p_j\right), \tag{5.17}$$

whose operator norm is at most $1/\tau$. Hence, by the bound on its Jacobian, the SoftMax is therefore $1/\tau$-Lipschitz on the bounded convex region explored during training:

$$\| S_\tau(a) - S_\tau(b) \|_2 \leq \frac{1}{\tau} \| a - b \|_2 . \tag{5.18}$$

Combining Eq. (5.8), (5.15) and (5.18) gives

$$\begin{aligned} \| p\left(\ell_{\text{num}}\right) - p\left(\ell_{\text{clean}}\right) \|_2 &= \| S_\tau\left(g\left(\ell_{\text{num}}\right)\right) - S_\tau\left(g\left(\ell_{\text{clean}}\right)\right) \|_2 \\ &\leq \frac{1}{\tau} \| g\left(\ell_{\text{num}}\right) - g\left(\ell_{\text{clean}}\right) \|_2 \\ &\leq \frac{L_g}{\tau} \| \ell_{\text{num}} - \ell_{\text{clean}} \|_2 \\ &\leq \frac{L_g}{\tau} \varepsilon . \end{aligned} \tag{5.19}$$

Thus, both the continuous sampling distribution and the discrete sampling probabilities depend smoothly on the PDE residual; with each module in the pipeline being Lipschitz, Theorem 2.2 ensures that their composition retains this regularity, providing a unified Lipschitz guarantee for the full sampling mechanism.

This theoretical analysis provides a clear explanation for the improved stability of GAN-based sampling compared with classical strategies such as residual-based adaptive refinement (RAR) or top- $k$ selection. Residual-based rules depend on ranking the entries of l , which is a discontinuous operation: arbitrarily small perturbations of the residual vector may cause non-small perturbations in the selected subset, making such methods inherently unstable under noise. In contrast, the GAN sampling pipeline is a composition of Lipschitz maps acting on a bounded-noise signal, yielding explicit global stability bounds.

From a more geometric viewpoint, the generator $G: Z \rightarrow \Omega$ continuously deforms an isotropic Gaussian measure on $Z$ into a structured distribution supported on a low-dimensional manifold $M \subset \Omega$. The continuity of the noise space Z and the infinite possibilities of sampling provide the source entropy for diversity:

$$z_i \neq z_j \Rightarrow G(z_i) \neq G(z_j) \,. \tag{6}$$

which prevents the sampler from collapsing onto a few fixed locations once a problematic region has been identified.

This contrasts sharply with classical residual-based adaptive refinement rules such as RAR or strict top-$k$ selection. Those strategies apply a *discontinuous* ranking or thresholding operator directly to the residual vector. Consequently, arbitrarily small perturbations in the residuals—for example, due to numerical noise or stochastic gradients—can lead to large, combinatorial changes in the chosen subset of points. In practice, this makes the sampling pattern highly sensitive to noise and can easily trigger local cycling, where the same small set of points is revisited again.

In our framework, by contrast, the discriminator and residual encoder act as a *smooth interface* between the PDE residual and the sampling distribution. The discriminator does not receive handcrafted labels; instead, labels are dynamically generated from the current residual field of the Phy-Transformer. After a short warm-up stage, we introduce a threshold τ that depends on the running mean residual.

Points with residuals above $\tau$ are labeled as "problematic", while the majority of points remain "normal". This deliberately sparse labeling emphasizes the exploration capability of GANs: the generator is attracted toward genuinely hard regions but still explores the surrounding domain due to the continuous nature of the latent space and the Lipschitz-stable mapping described above. During training, the discriminator is updated to minimize the standard binary cross-entropy:

$$L_D = -\mathbb{E}_{x\sim p_{\text{problematic}}} \log D(x) - \mathbb{E}_{z\sim p_z} \log\left(1 - D\left(G(z)\right)\right), \tag{7.1}$$

where $D(x)$ and $D\big(G(z)\big)$ are the discriminator's prediction probabilities for real points and generated points respectively. And the general loss of generator could also be set as:

$$L_G = -\mathbb{E}_{z\sim p_z} \log D\big(G(z)\big). \tag{7.2}$$

Concurrently, Phy-Transformer incorporates these generated points into its total loss as a weighted term, ,prioritizing areas with high discriminator scores or large residuals:

$$L_{\text{total}} = L_{\text{PDE}} + L_{\text{BC}} + L_{\text{IC}} + \lambda \cdot L_{\text{gen}}, \tag{7.3}$$

where $\lambda$ dynamically scales with the generator's confidence (i.e. $\lambda = \alpha * D\big(G(z)\big)$) to focus optimization on the critical spatiotemporal locations. To ensure effectiveness, only the points that satisfy the discriminator's screening (i.e. $D(G_i(z)) > \beta$ ) are allowed to be used for training. It should be noted that not only the generator and discriminator are trained alternately separately, but GAN is also alternately trained with Phy-Transformer. To facilitate a better understanding, the pseudocode for the overall training scheme is shown in Algorithm 1.

---

**Algorithm1** Proposed PhyTF-GAN framework for spatiotemporal PDEs

---

Generator *G*, Discriminator *D*
for j = 1 to N do
Generate Problematic Points:
    *S*=current PDE residuals+*Gaussian* noise
    $(x,t) \sim G(S)$ (sampling via Generator)

Update Discriminator:

Train $D$ to distinguish $\mathrm{D}_{\text{problematic}}$ vs. $\mathrm{D}_{\text{gen}}$

Loss: $L_D = -\left(\log D\left(D_{\text{problematic}}\right) + \log\left(1 - D\left(D_{\text{gen}}\right)\right)\right)$

$D \leftarrow D - \eta_D \nabla L_D$

Update Generator:

Reward: $R = -\log D(D_{\text{gen}})$ (discriminator score)

$G \leftarrow G - \eta_G \nabla R$

Update PINNs with Key Points:

Select high-priority points: $D_{\text{sel}} = \{(x,t) \mid D(G(S)) > \beta\}$

Loss: $L_{\text{total}} = L_{\text{PDE}}(D_{\text{sel}}) + L_{\text{IC/BC}} + L_{\text{PDE}}$

$u_\theta \leftarrow u_\theta - \eta \nabla L_{\text{total}}$

end for

RETURN: Trained PINNs model $u_\theta$, GAN components ($G$, $D$)

---

Considering the computational overhead incurred by training GAN, we also propose a faster training framework PhyTF-GAN-Skip, which reduces computational costs through a skip-step training approach. A hyperparameter $M$ is preset, and the GAN performs sampling once every $M$ steps. In the remaining iterations, the most recent sampling points from the previous round are selected for training. The pseudocode for the overall skip-step training strategy is shown in Algorithm 2.

---

**Algorithm 2** A skip-step training strategy for PhyTF-GAN

---

Generator $G$, Discriminator $D$, Phy-Transformer-GAN $P$

for j = 1 to N do

If j mod M = 1:

Generate Problematic Points:

$S$=current PDE residuals+*Gaussian* noise

$(x,t) \sim G(S)$ (sampling via Generator)

Update Discriminator:

Train $D$ to distinguish $\mathrm{D}_{\text{problematic}}$ vs. $\mathrm{D}_{\text{gen}}$

$D \leftarrow D - \eta_D \nabla L_D$

Update Generator:

$G \leftarrow G - \eta_G \nabla R$

end if

Update PINNs with Key Points:

$u_\theta \leftarrow u_\theta - \eta \nabla L_{\text{total}}$

end for

RETURN: Trained PINNs model $u_\theta$, GAN components ($G$, $D$)

### 2.3 Practical Considerations

As previously mentioned, the labels required for training are assigned through calculations by PINNs. The correctness of these labels will largely affect the overall accuracy of the network. If PINNs are given random initial weights during training, there is a high likelihood of generating incorrect labels, which can mislead the training of the GAN and ultimately cause the failure of PINNs. Therefore, before the formal training of the network, we will conduct a brief pretraining of PINNs to ensure that it can stably output labels for problematic regions without overfitting. This will effectively improve the stability of network training.

As mentioned in Eq. (2.1), the loss function of vanilla PINNs consists of three components, while Eq. (7.3) introduces an additional residual loss for problematic regions. Balancing these four loss components is complex. To simplify, we incorporate initial and boundary conditions as hard constraints into the network as shown in **Figure 5**, thereby avoiding the need to compute their contributions to the loss function. Furthermore, to enhance computational stability, we use finite difference methods instead of automatic differentiation for partial derivative calculations.

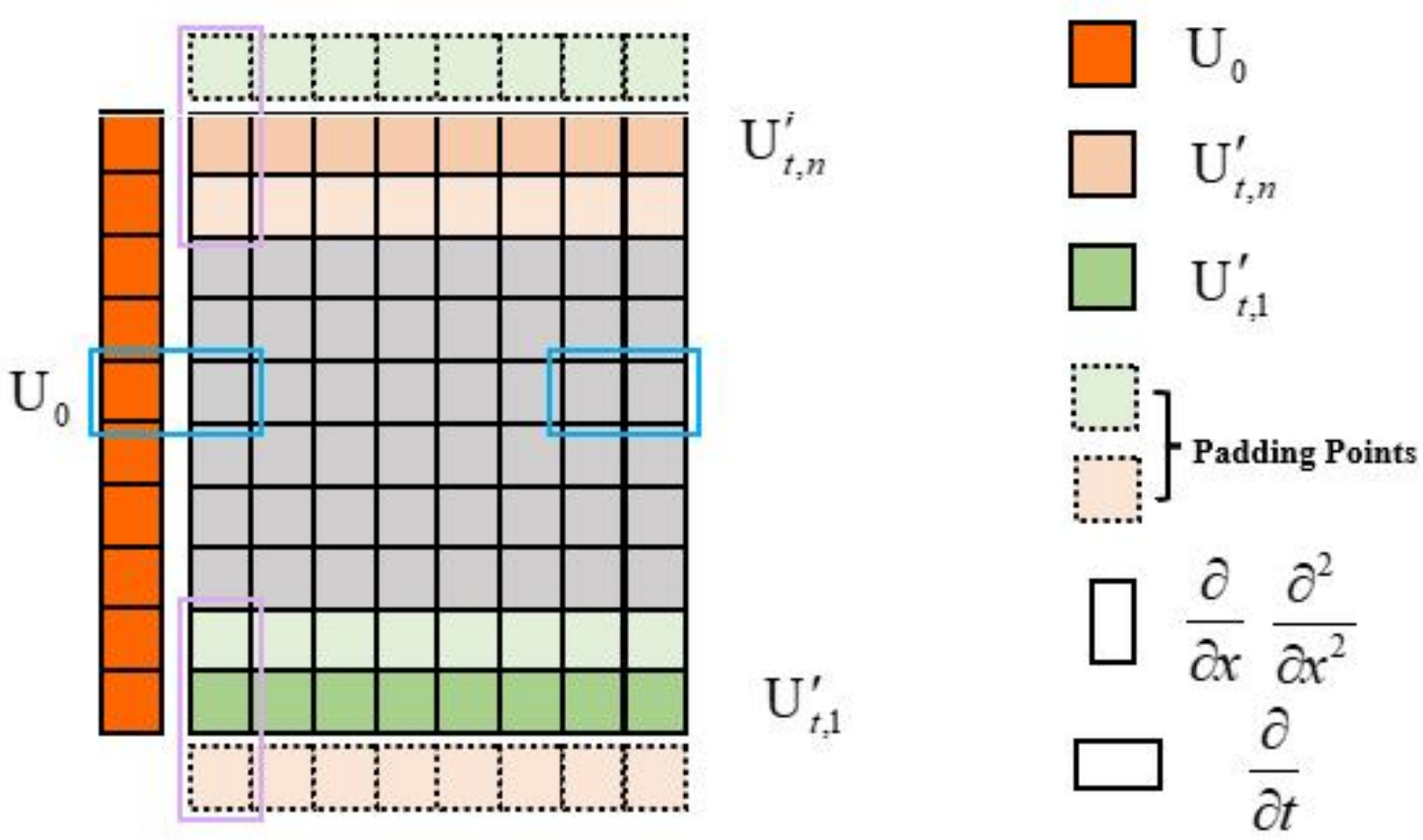

**Figure 5**. Discrete differentiation method with padding lattice.

Through finite difference methods, the derivative of variables can be shown as:

$$\frac{\partial u_{i,j}}{\partial x} = \frac{u_{i+1,j} - u_{i-1,j}}{2\Delta x}, \tag{8.1}$$

$$\frac{\partial u_{i,j}}{\partial y} = \frac{u_{i,j+1} - u_{i,j-1}}{2\Delta y}, \tag{8.2}$$

$$\frac{\partial^2 u_{i,j}}{\partial x^2} = \frac{u_{i+1,j} + u_{i-1,j} - 2u_{i,j}}{\Delta x^2}, \tag{8.3}$$

$$\frac{\partial^2 u_{i,j}}{\partial y^2} = \frac{u_{i,j+1} + u_{i,j\text{-}1} - 2u_{i,j}}{\Delta y^2}, \tag{8.4}$$

where $\Delta x$ and $\Delta y$ denote spatial separation distances of data. In traditional computing, this approach is widely adopted. By assigning a pre-defined filter to the convolutional network, the corresponding computational operator can be derived. The specific filter layers are as follows:

$$G_{Laplace} = \begin{bmatrix} 0 & 1 & 0 \\ 1 & -4 & 1 \\ 0 & 1 & 0 \end{bmatrix} \; G_x = \begin{bmatrix} -1 & 0 & 1 \\ -1 & 0 & 1 \\ -1 & 0 & 1 \end{bmatrix} \; G_y = \begin{bmatrix} 1 & 1 & 1 \\ 0 & 0 & 0 \\ -1 & -1 & -1 \end{bmatrix} \tag{9.1}$$

where $G_{Laplace}$ can be used as the Laplace operator for the variable, $G_x$ and $G_y$ can be used to calculate the partial derivatives with respect to space. In the same way, the derivative of the time term can be obtained. Take $G_t = [\text{-}1 \;\; 0 \;\; 1]$ and $\frac{\partial u}{\partial t}$ can be obtained as:

$$\frac{\partial u}{\partial t} = \frac{u_{t+1} - u_{t-1,}}{2\Delta t}, \tag{9.2}$$

and denotes the time distance.

**3 Numerical Experiment**

In this section, we present a comprehensive numerical evaluation of the proposed PhyTF-GAN framework. All experiments are implemented in PyTorch [58]. To ensure fair and consistent comparisons across different PDEs benchmarks, we employ the same network architecture and training protocol for all test cases. Model parameters are optimized using a two-stage strategy, where the Adam optimizer [59] is first applied to facilitate efficient exploration of the parameter space, followed by L-BFGS [60] for further refinement, with a total training budget of 10,000 iterations. All experiments are conducted on an NVIDIA RTX 4090 GPU.

The numerical studies are designed to assess the performance of PhyTF-GAN from multiple complementary perspectives, with an emphasis on both numerical accuracy and underlying training behavior. Specifically, we examine: **(1)** the sensitivity of the method to key hyperparameters and provide practical guidelines for their selection; **(2)** the robustness of the proposed framework to noise, by evaluating training stability and solution accuracy under perturbed residual signals and/or noisy supervision; **(3)** the evolution and spatiotemporal distribution of collocation points during training, which reveals how different sampling strategies adaptively focus on regions associated with complex solution features; **(4)** a frequency-domain analysis of solution errors, offering further insight into how different methods capture low and high-frequency components beyond conventional scalar error metrics; and **(5)** a set of representative case studies, in which PhyTF-GAN is compared with baseline approaches on selected PDE problems to illustrate its solution accuracy in practical settings.

### 3.1 Hyperparameter Sensitivity and Practical Selection Guidelines

The proposed PhyTF-GAN framework involves several algorithmic and training-related hyperparameters that play distinct roles in governing optimization behavior, sampling dynamics, and solution quality. A careful examination of these parameters is therefore essential for understanding the practical behavior of the method and for achieving reliable performance across different problem settings. In this subsection, we conduct a systematic investigation of key hyperparameters in PhyTF-GAN and provide empirical guidance for their selection.

Specifically, we examine: **(i)** the number of collocation points and their sampling frequency during training, which jointly control the resolution and temporal adaptivity of the sampling process; **(ii)** the threshold that governs the temporal progression of the causal training procedure; **(iii)** the screening parameter $\beta$, introduced to ensure the validity and effectiveness of GAN-generated collocation points by filtering low-quality samples; and **(iv)** the relative weights in the loss function, which balance data fidelity, physics constraints, and generative regularization.

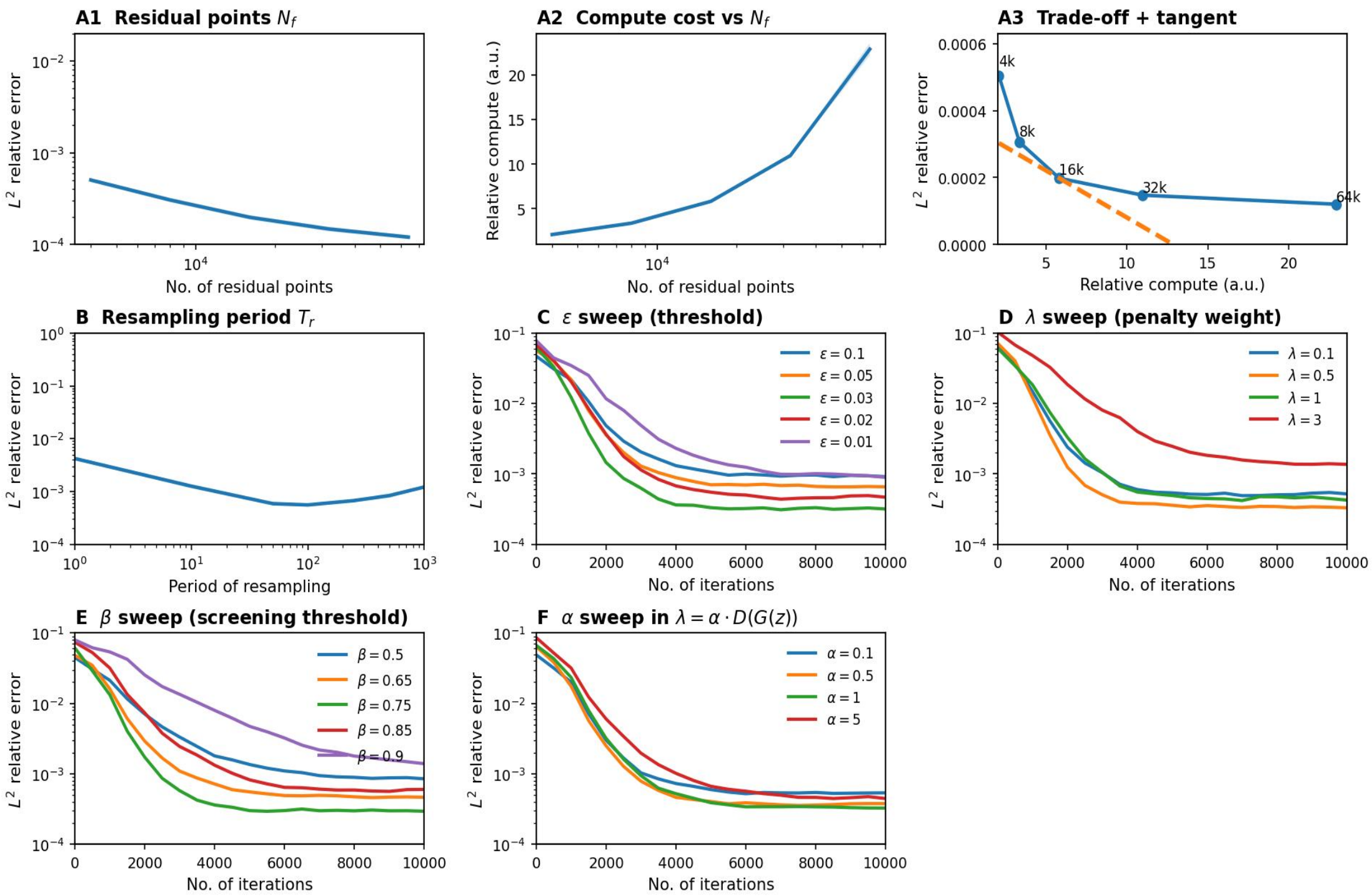

**Figure 6.** Sensitivity analysis of key hyperparameters in PhyTF-GAN.

**Figure 6** summarizes the sensitivity of PhyTF-GAN to several key hyperparameters related to sampling density, resampling dynamics, causal progression, and generative regularization. As shown in Figs. 6(A1–A3), increasing the number of residual collocation points $N_f$ consistently reduces the solution error but leads to a rapidly increasing computational cost. The combined accuracy–cost trade-off indicates a diminishing-returns regime beyond $N_f \approx 1.6\times10^4$, where further increases yield only marginal accuracy gains at substantially higher cost. Accordingly, $N_f = 16\,k$ is adopted in all subsequent experiments. The influence of the resampling period $T_r$ is shown in Fig. 6(B). Very frequent resampling provides limited additional benefit, while infrequent resampling degrades performance due to insufficient adaptation of the sampling distribution. The lowest error is achieved with an intermediate resampling interval, and we therefore resample every $T_r$=100 training iterations in all experiments. The threshold governing the temporal progression of causal training (Fig. 6(C)) controls a trade-off between convergence speed and final accuracy. Here, ε specifies a percentile-based criterion under which a time step is regarded as "solved", i.e., a time step is marked as solved only when its associated loss falls within the lowest ε fraction of losses over all time steps. Thresholds that are too large result in premature temporal advancement and higher terminal errors, whereas excessively small thresholds slow down convergence. Based on the observed balance between accuracy and training efficiency, we select $\varepsilon = 0.03$ as the default value. A similar trade-off is observed for the penalty weight $\lambda$ (Fig. 6(D)), where moderate values lead to faster convergence and lower final errors. Here, the causal penalty term is first normalized to have a comparable magnitude to the primary PDE residual loss, so that $\lambda$ primarily controls the relative emphasis rather than compensating for scale mismatch. Accordingly, in all subsequent experiments, we set $\lambda = 0.5$. The effects of generative regularization are illustrated in Figs. 6(E) and 6(F). The screening threshold $\beta$ has a pronounced impact on performance: intermediate values effectively filter low-quality GAN-generated collocation points while preserving sampling diversity. Based on Fig. 6(E), we choose $\beta = 0.75$. In contrast, the scaling parameter $\alpha$ in the discriminator-based weighting exhibits relatively mild sensitivity over a broad range; we therefore adopt $\alpha$=1 as a stable default. With these choices, all subsequent numerical experiments are conducted using a fixed and well-justified hyperparameter configuration, ensuring both reproducibility and a favorable balance between accuracy, efficiency, and training stability.

### 3.2 Sensitivity to Noisy Sampling Scores

In this subsection, we investigate the robustness of different sampling strategies to noise in the sampling score, rather than noise in the PDE training objective itself. For each candidate collocation point $x$, a clean scalar score $s(x)$ is computed from the physics residual and is used only for ranking or selecting points during adaptive or generative resampling. To emulate uncertainty in residual-based scoring—arising, for example, from numerical differentiation, stochastic optimization, or early-stage training instability—we perturb the score used for point selection and define an observed score

$$\tilde{s}(x) = s(x) + \sigma\varepsilon(x), \varepsilon \sim N(0,1), \tag{10}$$

where $\sigma$ controls the noise level. Importantly, the PDE loss used for training and backpropagation remains unchanged and is always computed from the clean residual. As a result, this experiment isolates the sensitivity of the sampling mechanism to score uncertainty, without altering the underlying optimization objective. Since PhyTF-GAN requires a pretraining stage, all models involved in our experiments are initialized from a checkpoint obtained after 100 epochs of pretraining.

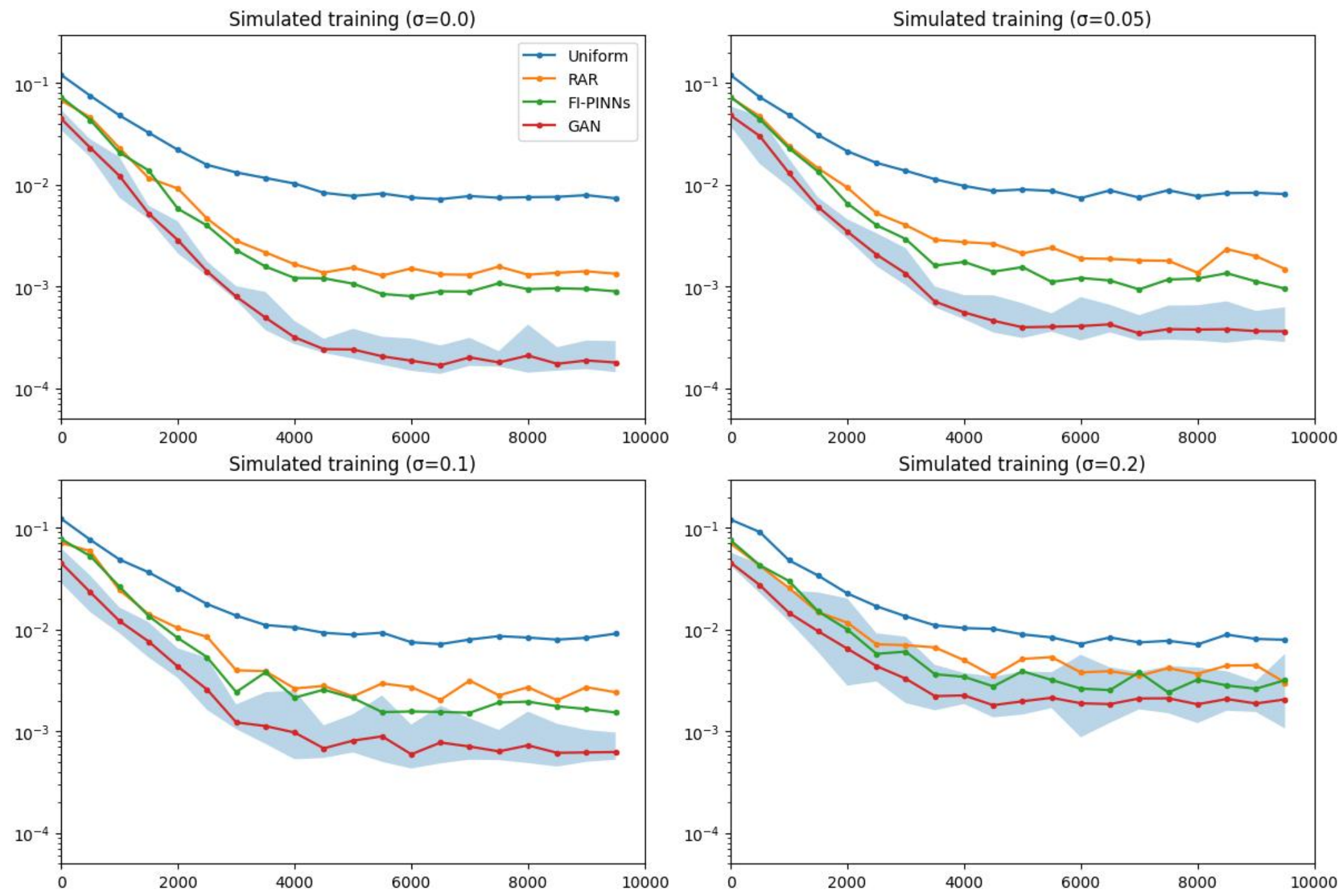

**Figure 7**. Robustness of different sampling strategies under noisy residual scores.

The training curves in **Figure 7** illustrate how different methods respond to increasing noise levels. Uniform sampling, which does not rely on residual-based scores, remains largely insensitive to $\sigma$, serving as a reference baseline. In contrast, methods that depend on score-based ranking exhibit progressively degraded convergence as the noise level increases, reflecting the accumulation of mis-ranked collocation points during resampling. Compared with residual-driven adaptive baselines, PhyTF-GAN demonstrates a more graceful degradation under noisy scores, maintaining faster convergence and lower terminal error across all noise levels.

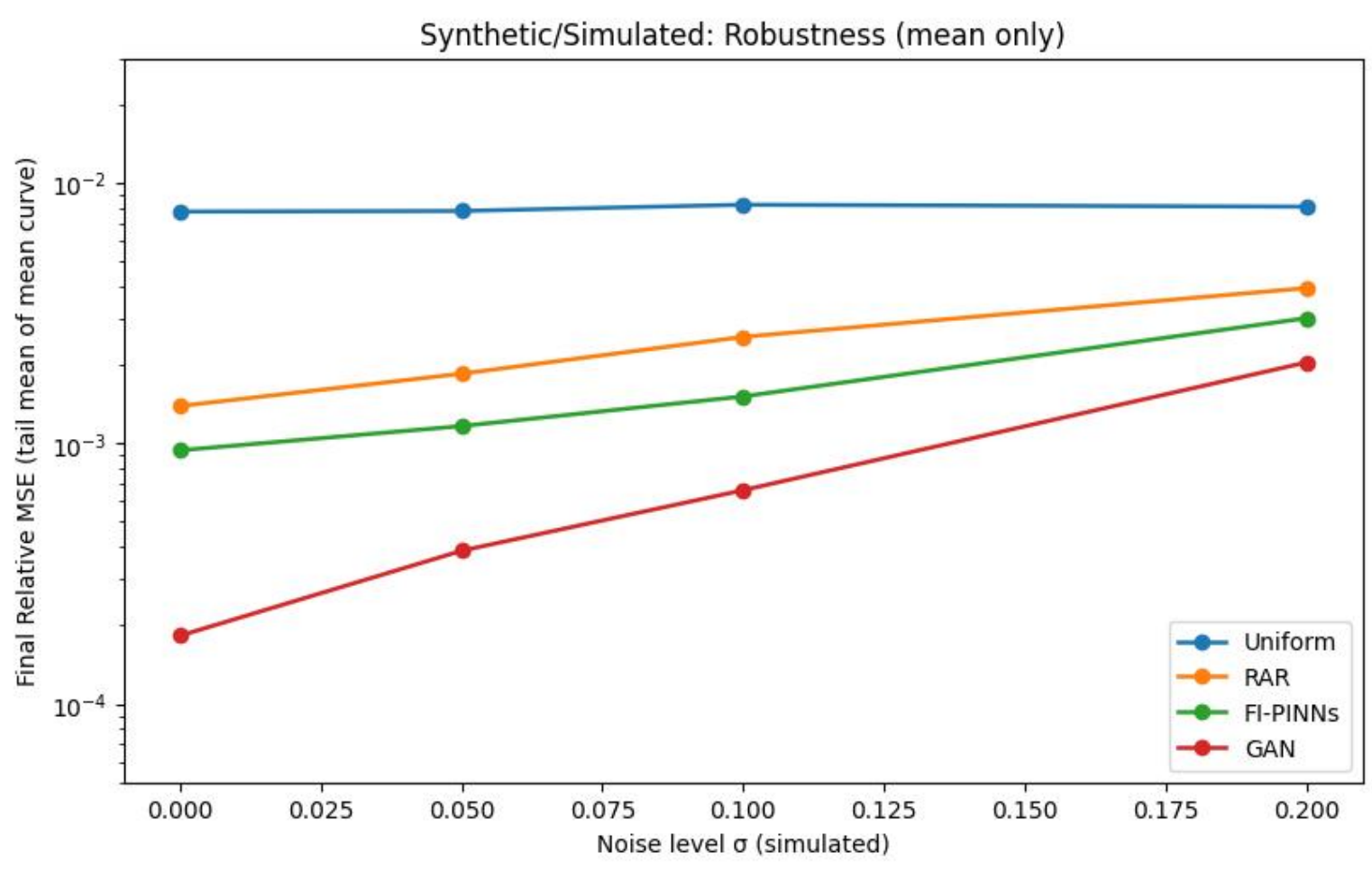

**Figure 8**. Quantitative comparison of accuracy to noisy residual scores.

The quantitative summary in **Figure 8** further confirms this trend by reporting the tail-averaged error as a function of $\sigma$. While the error of score-based baselines increases monotonically with noise intensity, PhyTF-GAN consistently achieves the lowest error. These results indicate that the proposed generative sampling strategy is more robust to uncertainty in residual-based scoring, which is critical in practice where sampling decisions are often made based on imperfect or noisy indicators rather than exact residual values.

### 3.3 Evolution and Distribution of Sampling Points

To better understand how the proposed method allocates collocation points during training, we investigate the evolution of sampling distributions together with the corresponding solution fields. We first visualize the temporal evolution of sampling points and solution quality for several representative PDEs, illustrating how sampling, prediction, and local errors evolve jointly. We then provide a direct comparison of sampling distributions across different methods by inspecting the first three sampling updates under identical settings. This combined analysis aims to clarify not only where points are sampled, but also how different sampling mechanisms adapt over time.

We begin with a Poisson Equation to illustrate how the sampling distribution produced by the proposed method evolves during training. For each stage, we report four components: the spatial distribution of collocation points, the predicted solution field, a one-dimensional slice comparing the prediction with the exact solution, and the histogram of pointwise errors (see Figure 9). At Stage 1, sampling points are relatively diffuse and only weakly correlated with the underlying solution structure, leading to visible discrepancies in the predicted field and noticeable bias in the one-dimensional profile. As training proceeds from Stage 1 to Stage 3, sampling points progressively concentrate in regions associated with larger solution variation, which coincides with a gradual alignment between the predicted and exact solutions. This improvement is reflected not only in the field visualization but also in the narrowing of the error histogram, indicating a systematic reduction of large-error samples. These results show that the sampling distribution evolves coherently with the solution accuracy rather than remaining static.

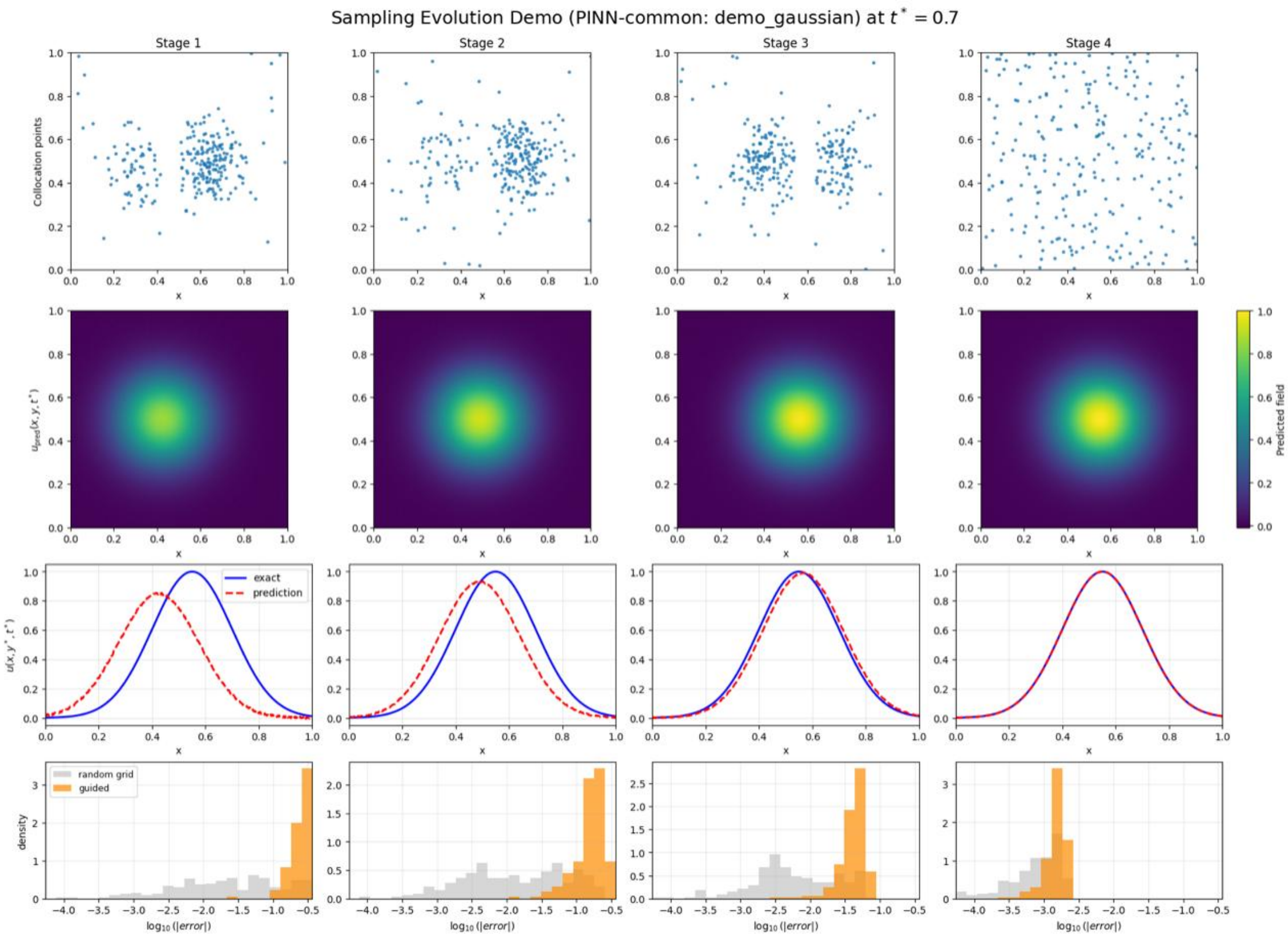

**Figure 9**. Sampling Evolution, Predicted Field, 1D Profile and Tail-Shift Histogram of a Poisson Equation.

Next, we consider a Burgers Equation with shock formation to examine how the sampling strategy adapts to localized, non-smooth solution features. In the early stage, as shown in the **Figure 10**, sampling points are sparsely distributed near the shock region, resulting in a smeared transition in the predicted profile. As training advances, the sampling distribution increasingly concentrates along the shock trajectory, allowing the model to better resolve the sharp gradient. This concentration is accompanied by a marked improvement in the predicted one-dimensional profile, where the discontinuity becomes sharper and more accurately located. The corresponding error histograms further demonstrate that high-error samples near the shock are progressively reduced, highlighting the effectiveness of adaptive sampling in capturing localized, non-smooth solution features.

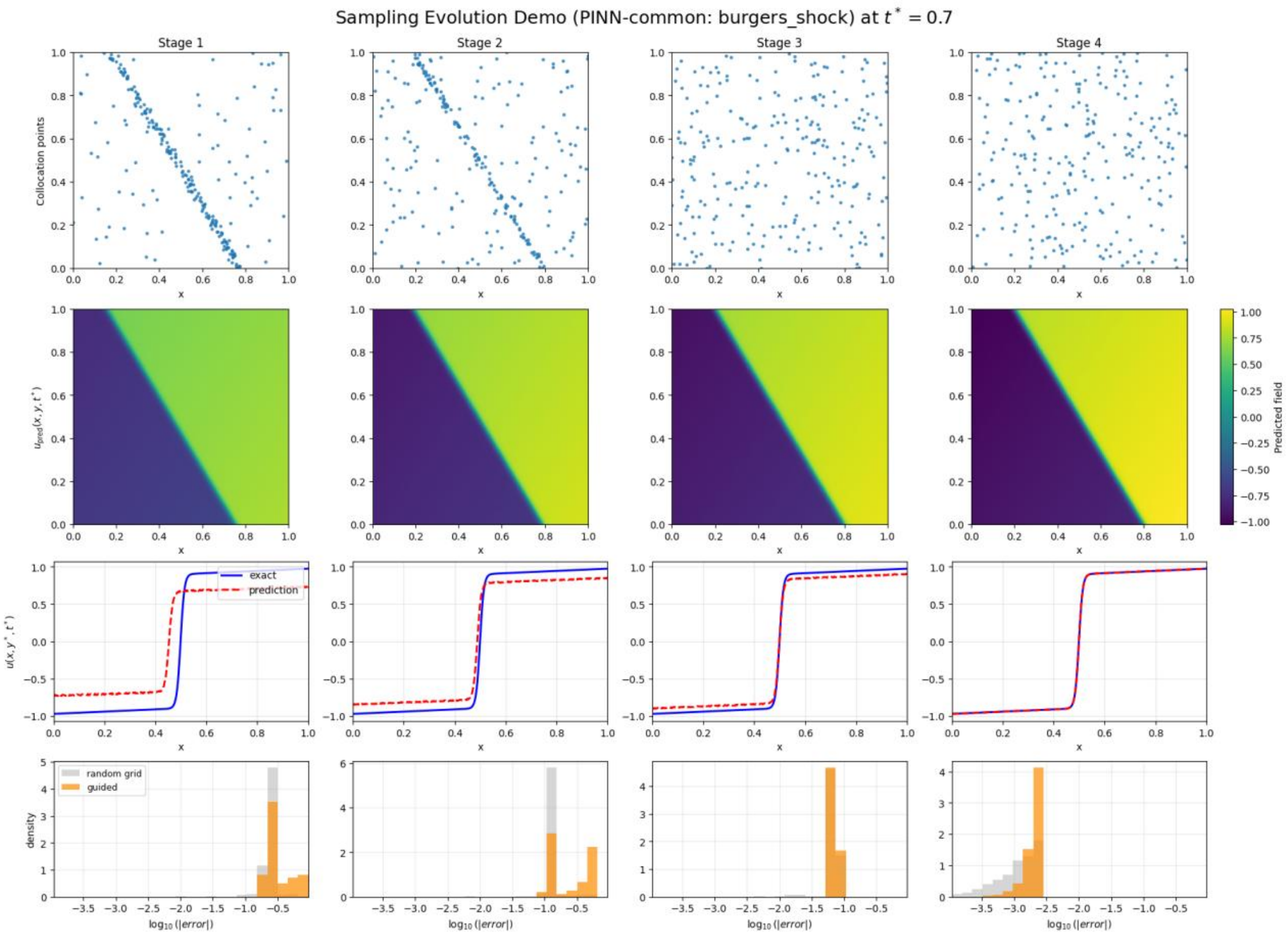

**Figure 10**. Sampling Evolution, Predicted Field, 1D Profile and Tail-Shift Histogram of a Burgers Equation.

Finally, we study a Helmholtz Equation characterized by highly oscillatory solutions to assess the ability of the sampling strategy to capture fine-scale structures. In this case, uniform or weakly adaptive sampling is insufficient to capture the fine-scale structure of the solution. As shown in **Figure 11**, the visualizations indicate that, as training progresses, sampling points gradually adapt to the oscillatory patterns, leading to improved phase alignment and amplitude accuracy in the predicted profiles. Compared with earlier stages, later stages exhibit a closer match to the exact solution across the entire domain, as well as a noticeable shift of the error distribution toward lower magnitudes. This example demonstrates that the proposed sampling strategy can adapt not only to localized singularities but also to globally oscillatory structures. Taken together, these results indicate that the proposed method exhibits robust adaptive behavior across PDEs with fundamentally different solution characteristics, including smooth fields, sharp gradients, and high-frequency oscillations.

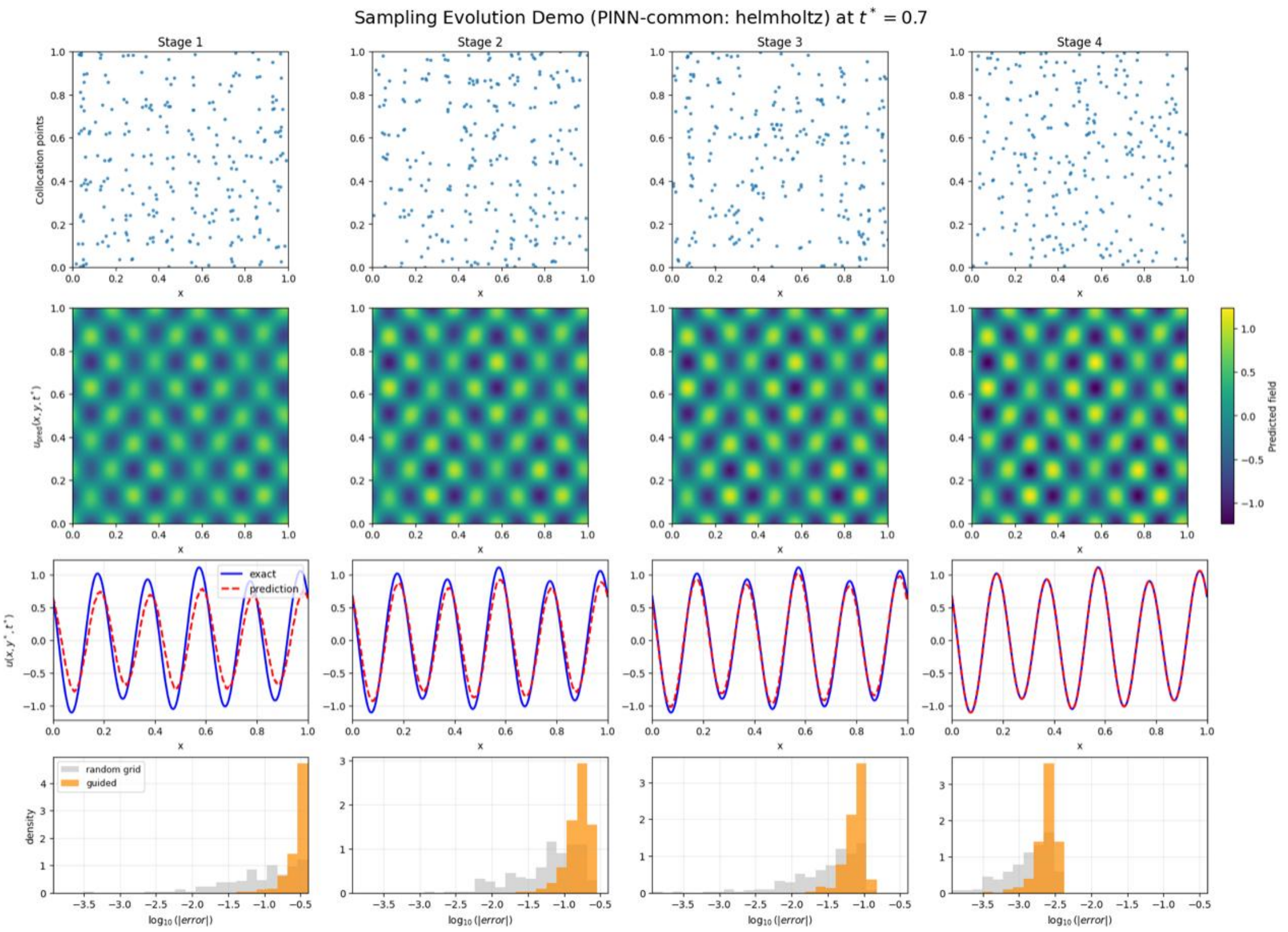

**Figure 11**. Sampling Evolution, Predicted Field, 1D Profile and Tail-Shift Histogram of a Helmholtz Equation.

While the above results highlight the temporal evolution of sampling for the proposed method on individual PDEs, **Figure 12** and **Figure 13** provide a direct comparison of sampling distributions across different methods. Specifically, we visualize the collocation points obtained during the first three sampling updates for uniform sampling, RAR, FI-PINN, AAS-PINN, and the proposed GAN-based approach. Uniform sampling remains spatially homogeneous across all updates and does not adapt to solution features. RAR gradually increases point density in high-residual regions, but the resulting distributions remain scattered, and exhibit delayed structural alignment. FI-PINN produces highly structured sampling patterns at early stages, often concentrating points along narrow manifolds, which may lead to insufficient coverage in other regions. AAS-PINN improves adaptability by adjusting the sampling distribution over successive updates, yet its control over the overall distribution shape remains limited.

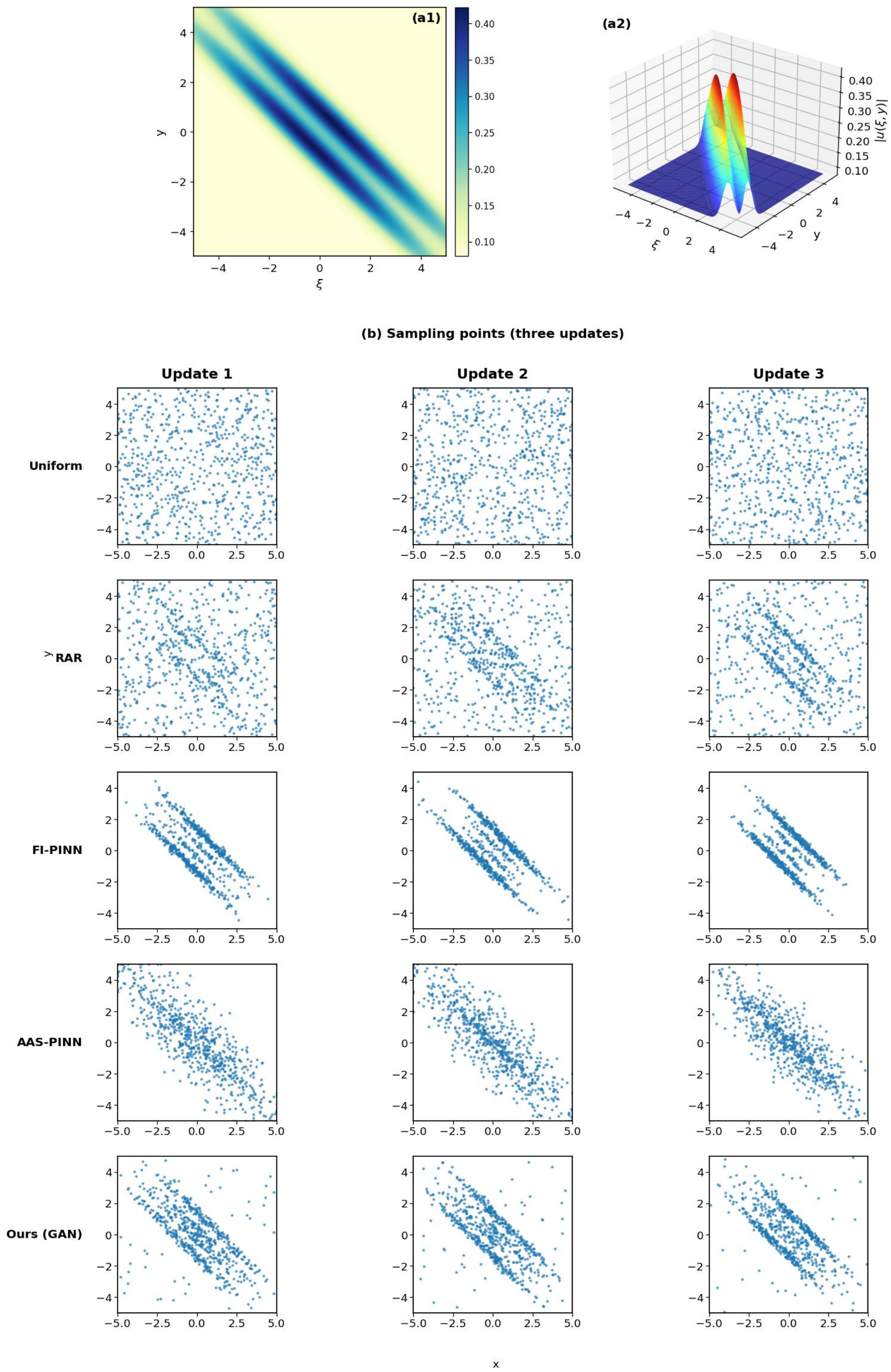

**Figure 12**. Comparison of sampling distributions across different adaptive sampling strategies on Fisher–KPP equation.

In contrast, the proposed GAN-based method consistently generates sampling distributions that align with the underlying solution structures from the earliest updates while maintaining sufficient global coverage. Across successive updates, the sampling patterns evolve smoothly and remain stable, reflecting a distribution-level adaptation rather than pointwise ranking alone. These results demonstrate that the proposed method achieves a favorable balance between concentration and coverage, enabling more effective and robust sampling compared with existing adaptive strategies.

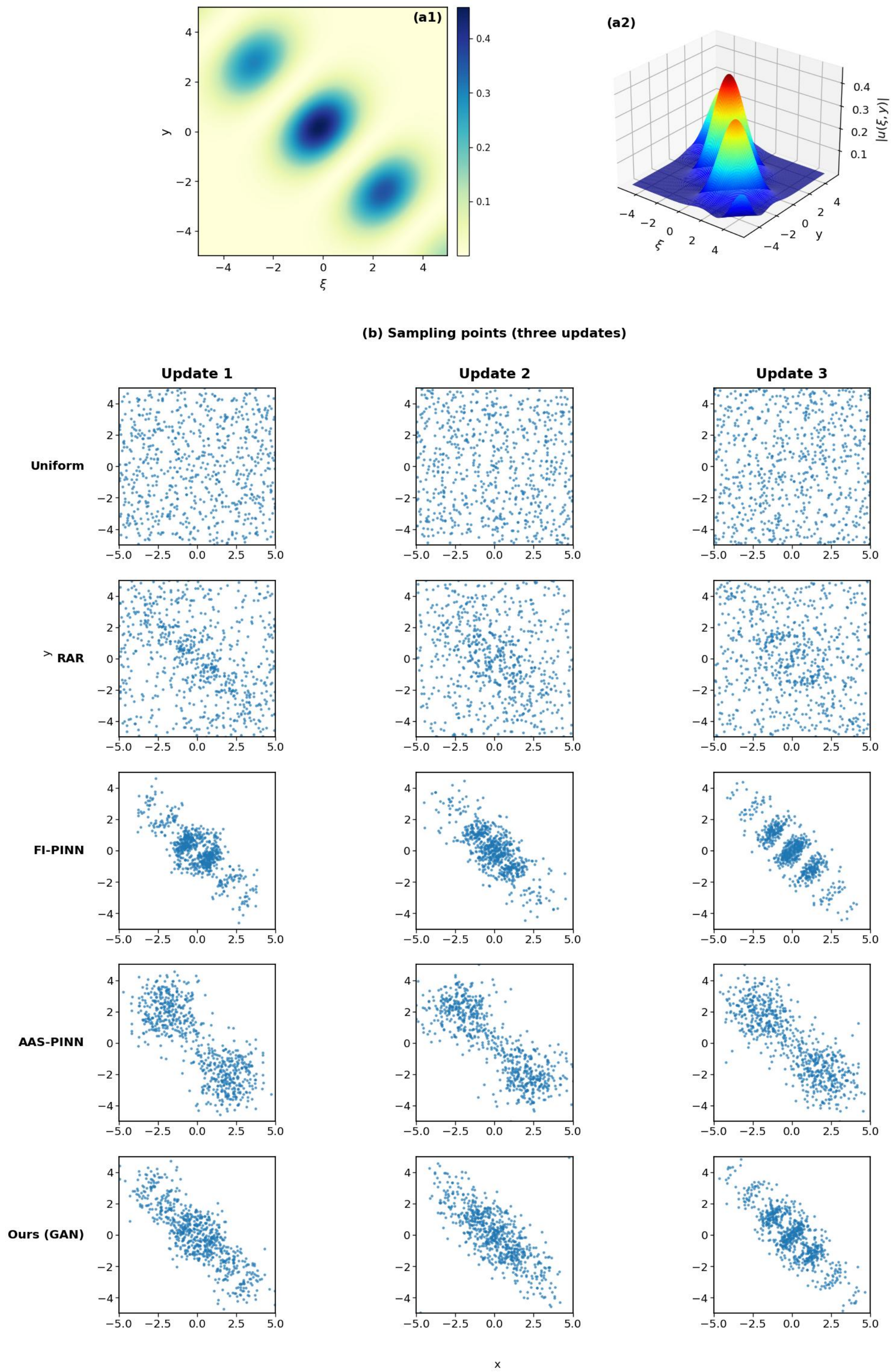

**Figure 13**. Comparison of sampling distributions across different adaptive sampling

strategies on Nonlinear Schrödinger equation.

### 3.4 Frequency-Domain Error Analysis

To further investigate how different sampling strategies influence the frequency content of the learned solutions, we conduct a frequency-domain error analysis on a one-dimensional bistable reaction–diffusion equation whose solution is dominated by a narrow internal layer. This class of problems is particularly suitable for frequency analysis, as the sharp transition region concentrates high-frequency components and poses a well-known challenge for physics-informed learning methods.

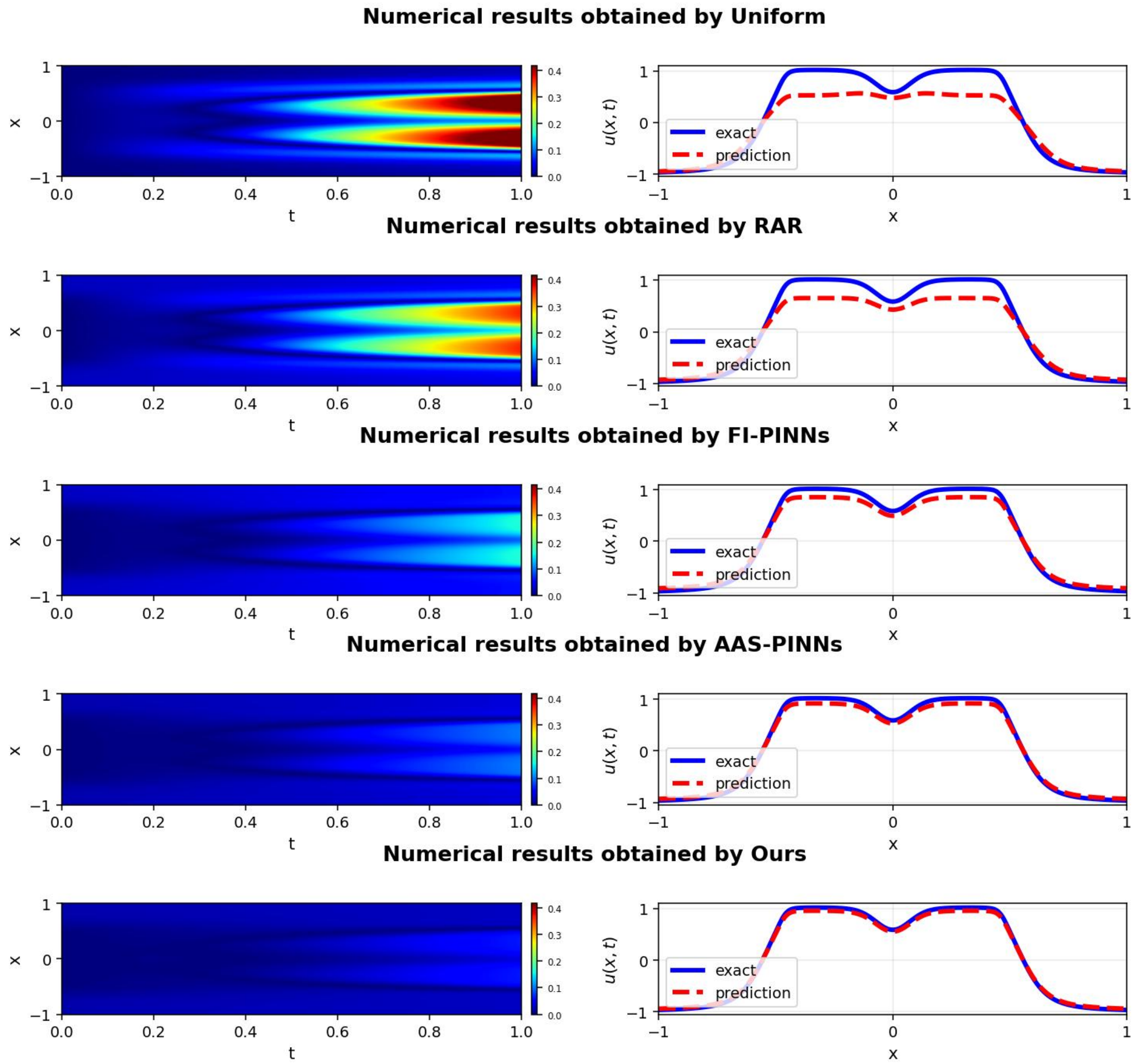

**Figure 14**. Numerical solutions of a one-dimensional bistable reaction–diffusion equation with a sharp internal layer.

**Figure 14** first compares the numerical solutions obtained by uniform sampling, RAR, FI-PINNs, AAS-PINNs, and the proposed method. The spatiotemporal solution fields and the one-dimensional slices at the fixed time show that uniform sampling leads to excessive smoothing near the internal layer, resulting in a diffused transition region. Residual-based adaptive methods improve the overall solution accuracy, but noticeable interface smearing remains. FI-PINNs and AAS-PINNs exhibit intermediate behavior between uniform sampling and the proposed method: FI-PINNs partially sharpen the transition layer compared with uniform sampling, yet still display small-amplitude oscillations and residual interface smearing in the one-dimensional profile, indicating the presence of unresolved high-frequency errors. AAS-PINNs further enhance the interface resolution by adaptively adjusting the sampling distribution; however, oscillatory artifacts persist near the transition region, suggesting that high-frequency components are not fully suppressed. In contrast, the proposed method captures a sharper and more accurate transition layer, yielding closer agreement with the reference solution in both the spatiotemporal domain and the one-dimensional profile. Overall, these observations indicate that while FI-PINNs and AAS-PINNs improve local resolution, they remain sensitive to high-frequency residuals concentrated near the internal layer, whereas the proposed method achieves a more balanced representation without introducing non-physical oscillations.

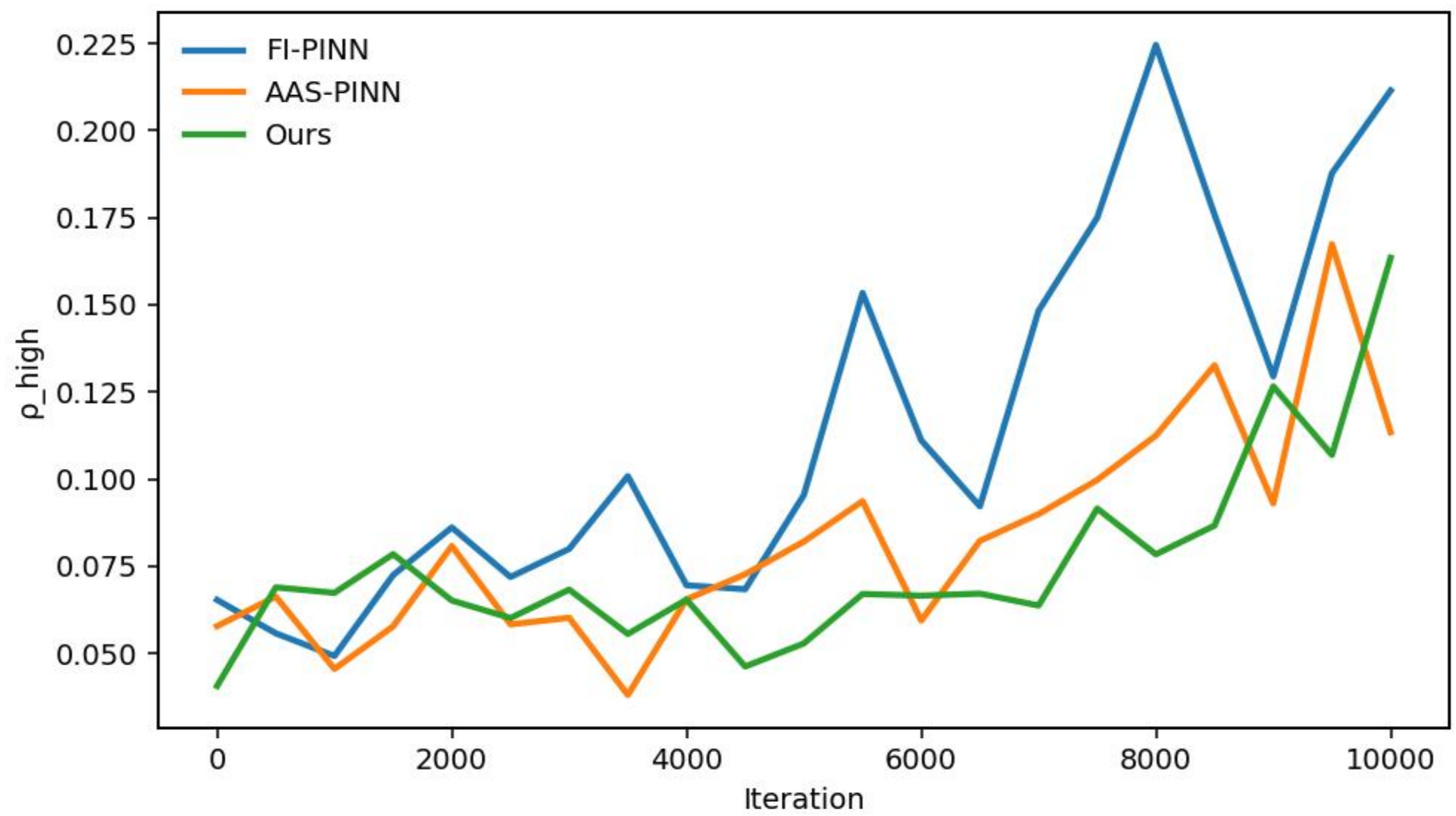

**Figure 15**. Evolution of the high-frequency energy ratio $\rho_{\text{high}}$ of the solution error during training.

While these solution profiles reveal clear qualitative differences among sampling strategies near the internal layer, they do not fully explain the underlying frequency mechanisms responsible for such behavior. To make this mechanism explicit, we further analyze the frequency composition of the error during training. Herein, we briefly define the quantities involved. Let

$$e(x,t) = u_{\theta}(x,t) - u(x,t) \tag{11.1}$$

denote the solution error, where $u_\theta$ is the network prediction and $u$ is the reference solution. For a fixed time snapshot, we compute the spatial Fourier transform $\hat{e}(k)$ of $e(x, t)$ and define the corresponding energy spectrum as

$$E(k) = |\,\hat{e}(k)\,|^{2}. \tag{11.2}$$

For two-dimensional problems, $E(k)$ is further averaged over circular shells in Fourier space to obtain a radial energy spectrum as a function of the wavenumber magnitude $k$. Based on this spectrum, we define the high-frequency energy ratio

$$\rho_{\text{high}} = \frac{\int_{k>k_c} E(k)\ dk}{\int E(k)\, dk}, \tag{11.3}$$

where $k_c$ is a fixed cutoff separating low- and high-frequency modes. This quantity measures the relative contribution of fine-scale (high-frequency) error components during training. **Figure 15** reports the ratio of high-frequency energy in the error (or residual-based loss) throughout optimization. A common trend across methods is that this high-frequency ratio gradually increases as training proceeds, consistent with the fact that the solution is governed by a narrow internal layer whose accurate resolution requires capturing fine-scale structures.

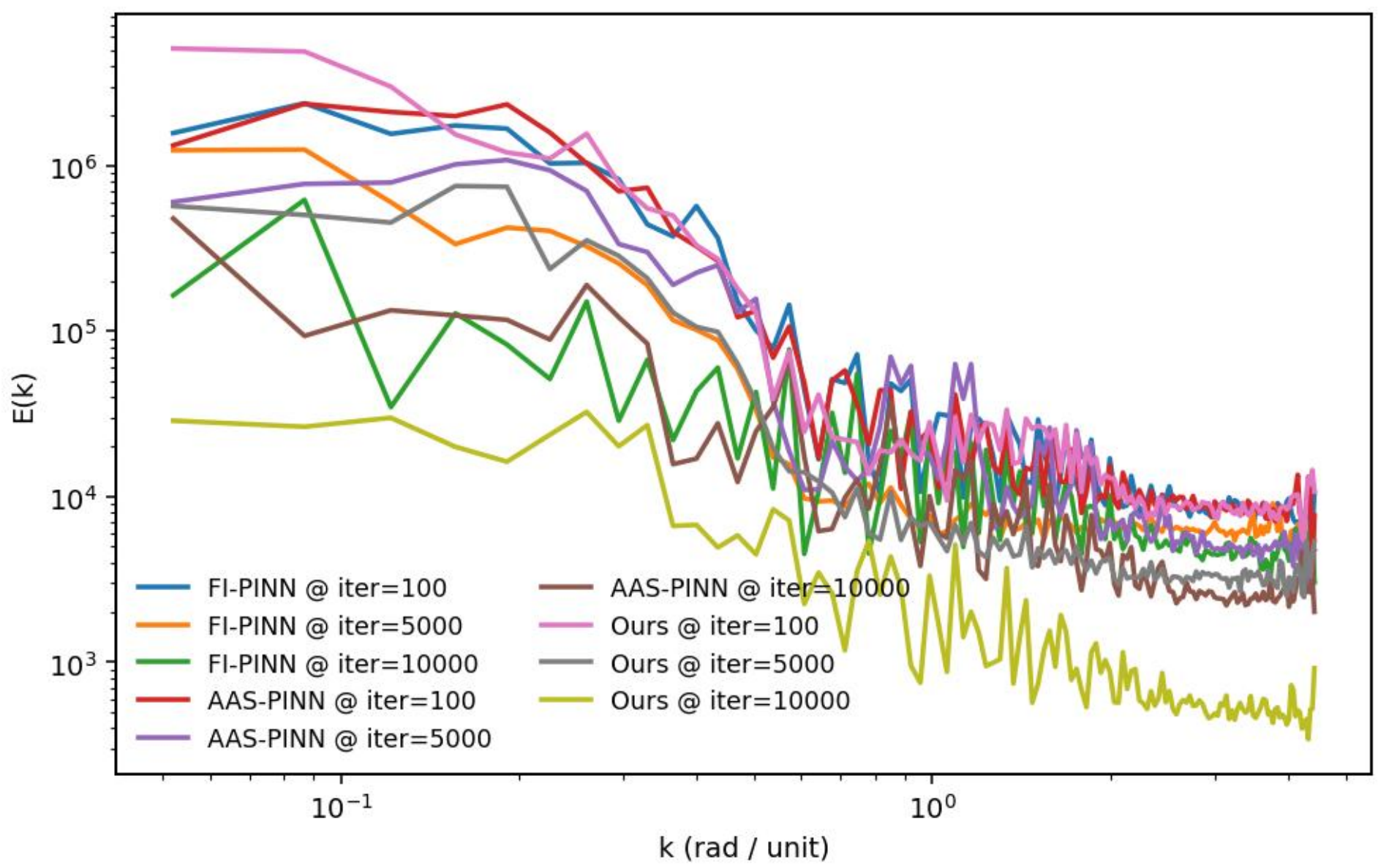

**Figure 16**. Radial energy spectrum $E(k)$ of the solution error at selected training checkpoints.

More detailed differences emerge from **Figure 16**, which presents the evolution of the energy spectrum $E(k)$ at representative training checkpoints. Although all methods retain non-negligible energy in the high-wavenumber range, FI-PINN and AAS-PINN exhibit relatively slower decay at large $k$, indicating persistent high-frequency error content. In contrast, our proposed method shows a clearer attenuation of energy in the high-frequency regime at later iterations, reflecting more effective suppression of fine-scale numerical errors. Overall, although high-frequency components become increasingly relevant during training for all methods, the proposed approach exhibits a more favorable attenuation of high-frequency energy at later stages, which is consistent with the sharper and more accurate solution profiles observed in **Figure 14**.

### 3.5 Case Studies

This subsection presents a set of case studies involving several representative partial differential equations. The goal is to complement the preceding systematic analyses by examining the proposed method in concrete problem settings. In all numerical experiments, we use a fixed training configuration across different PDEs to ensure fair comparison. Specifically, unless otherwise stated, we employ a Transformer-based surrogate model with a fixed architecture across all experiments. The Transformer consists of Decoder layers with a model dimension of 128 and 8 attention heads and is trained using the Adam optimizer with a learning rate of $1\times10^{-3}$. All models are trained for 10,000 iterations. Key hyperparameters in the proposed method are fixed throughout all experiments, including the causal penalty weight $\lambda = 0.5$, the temporal progression threshold $\varepsilon = 0.03$ (defined as a relative criterion based on the ranking of per-time-step losses), and the GAN-related sampling parameters. In particular, we use $N_f = 16{,}000$ residual collocation points, a resampling period of $T_r = 100$ iterations, a screening threshold $\beta = 0.75$, and a discriminator scaling factor $\alpha = 1$. These values are selected based on the sensitivity analysis and are kept unchanged in all subsequent experiments unless explicitly noted. For each equation, the solution is investigated under different parameter regimes, allowing us to assess how variations in problem stiffness, nonlinearity, or scale separation influence the learning process. In addition, targeted ablation experiments are performed to isolate the effects of key components within the proposed framework and to evaluate their contributions under varying conditions. These case studies provide a more detailed and intuitive perspective on the practical behavior of the proposed method across diverse PDE scenarios

### 3.5.1 Allen-Cahn Equation

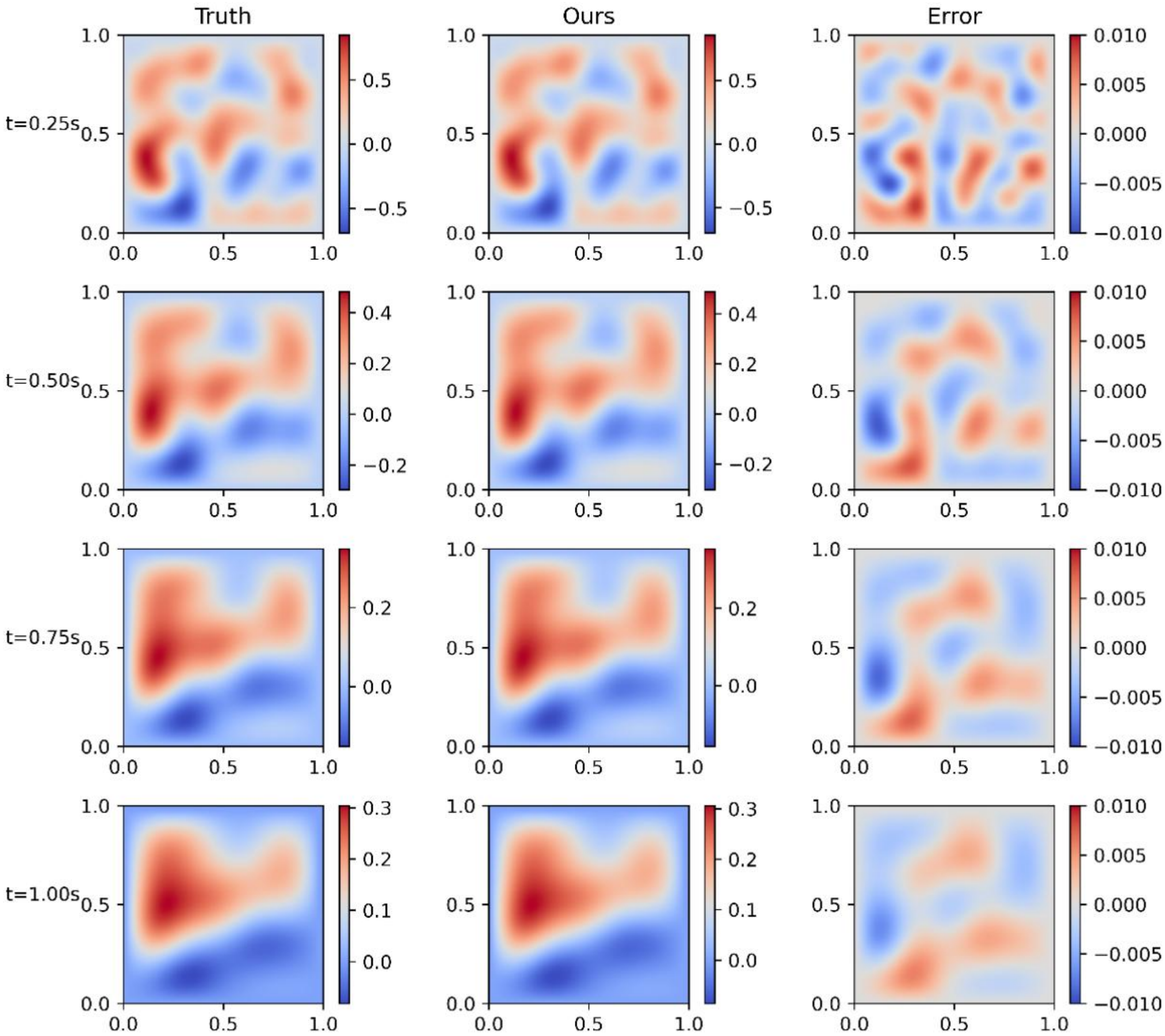

**Figure 17**. Results of Allen-Cahn Equation when $\varphi = 1$.

Our first example is the Allen-Cahn Equation, which conventional PINN models are known to struggle with. The general form of the Allen-Cahn Equation could be seen as:

$$\frac{\partial c}{\partial t} = -\frac{c^3 - c}{\varepsilon^2} + \varphi\left(\frac{\partial c^2}{\partial x^2} + \frac{\partial c^2}{\partial y^2}\right) \tag{12}$$

where $c$ can be used to describe two different phases of a material, and $\varphi$ denotes interface width parameter, which controls the width of the transition region between two different phase states. Herein, we consider three interface width parameters, $\varepsilon = 0.5, 1, 2$, to systematically examine the impact of interface thickness on the numerical behavior of the Allen–Cahn equation. For clarity of presentation, **Figure 17** focuses on the representative case $\varepsilon = 1$ and reports the corresponding concentration field $c$. Specifically, the figure displays the ground-truth solution, the prediction produced by the proposed PhyTF-GAN framework, and the absolute error between them at four representative time instances: $t$ = 0.25s, $t$ = 0.50s, $t$ = 0.75s and $t = 1\text{s}$.

### 3.5.2 Klein-Gordon Equation

The next example we consider is the Klein-Gordon Equation, and we hope it can further demonstrate the effectiveness of our framework. As a fundamental equation in relativistic quantum mechanics, it could be used to describe relativistic spin-0 particles such as the pion. The general form of the Klein-Gordon Equation could be seen as:

$$\begin{gathered}\frac{\partial^2 u}{\partial t^2}(x,y,t) = \nabla^2 u(x,y,t) - m^2 u(x,y,t) \\ u(x,y,t) = 0, \forall (x,y) \in \partial\Omega, t \geq 0\end{gathered} \tag{13}$$

where $\nabla^2 = \frac{\partial^2}{\partial x^2} + \frac{\partial^2}{\partial y^2}$, $u$ and $m$ represent the scalar field and the mass of the particle respectively. Herein, we consider three representative mass parameters, $m$ = 1, 3, 5, to systematically investigate the influence of mass on the dynamical behavior of the Klein–Gordon equation. Larger values of $m$ induce higher-frequency temporal oscillations in the scalar field, thereby increasing the learning difficulty for physics-informed models. Among these settings, we select $m = 3$ as a representative intermediate case for detailed visualization and discussion. **Figure 18** reports the numerical results of the scalar field $u$ under this setting, including the ground-truth solution, the predicted solution obtained by the proposed framework, and the corresponding absolute error. These quantities are presented at four representative time instances: $t = 1\text{s}$, $t = 2\text{s}$, $t = 4\text{s}$ and $t = 8\text{s}$, which capture different stages of the oscillatory evolution of the solution. The close agreement between the predicted and reference solutions across these time instances further demonstrates the reliability of the proposed method in resolving oscillatory wave dynamics.

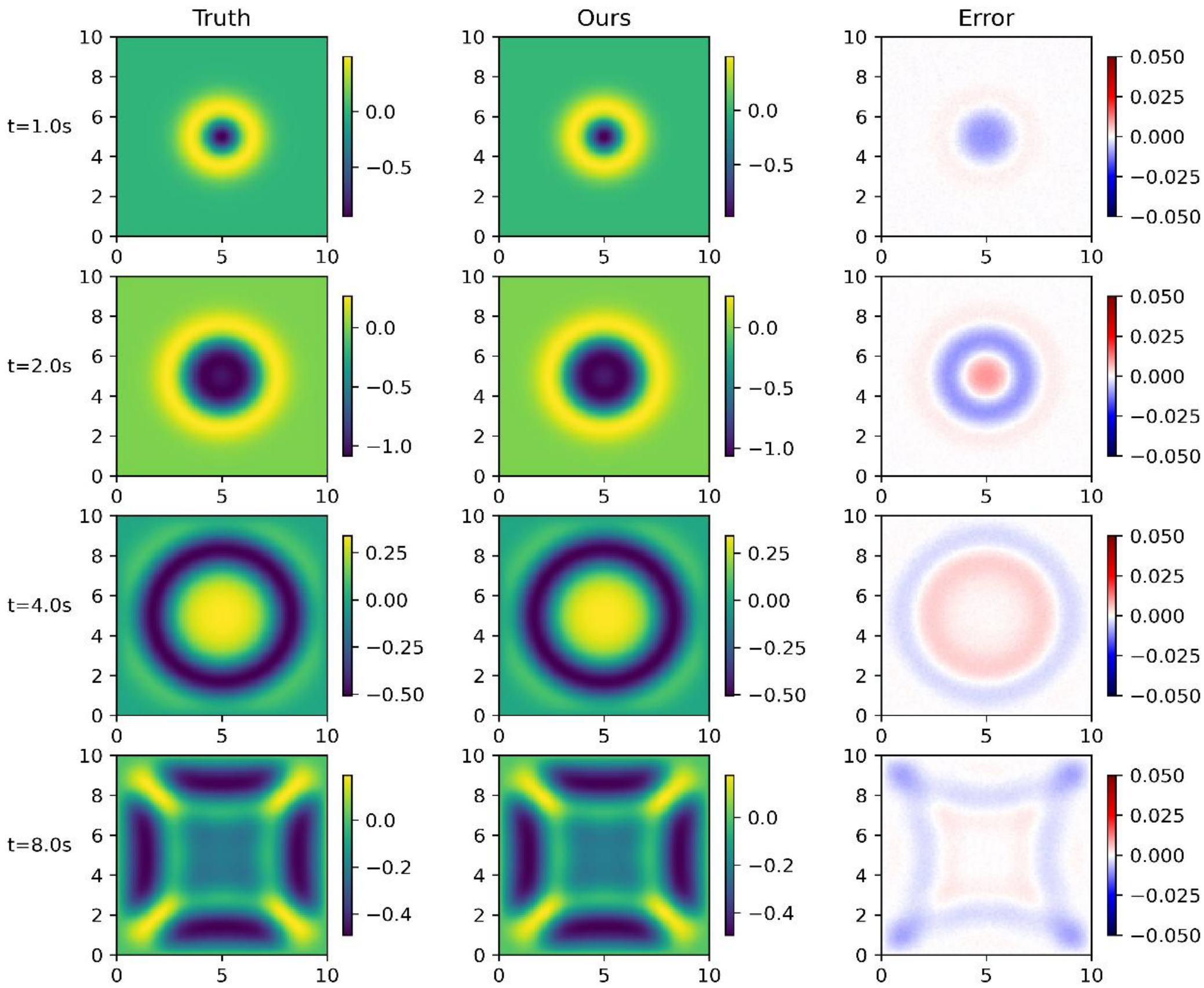

**Figure 18**. Results of Klein Gordon Equation when $m$ = 3.

### 3.5.3 Navier-Stokes Equation

For our last example, we consider the Navier-Stokes Equation, a core system of partial differential equations describing the motion of Newtonian fluids. Derived from the laws of conservation of mass, momentum, and energy, they emphasize the influence of viscous forces on fluid motion. In this section, we take into account a classical two-dimensional decaying Navier-Stokes Equation and the general form of it is as follows:

$$\begin{gathered}\frac{\partial u}{\partial t}+u\frac{\partial u}{\partial x}+v\frac{\partial u}{\partial y}=-\frac{1}{\rho}\frac{\partial p}{\partial x}+\frac{v}{Re}\left(\frac{\partial^2 u}{\partial x^2}+\frac{\partial^2 u}{\partial y^2}\right),\\ \frac{\partial v}{\partial t}+u\frac{\partial v}{\partial x}+v\frac{\partial v}{\partial y}=-\frac{1}{\rho}\frac{\partial p}{\partial y}+\frac{v}{Re}\left(\frac{\partial^2 v}{\partial x^2}+\frac{\partial^2 v}{\partial y^2}\right),\\ \frac{\partial u}{\partial x}+\frac{\partial v}{\partial y}=0, B(u,v,p)=0,(x,y)\in\Gamma,\end{gathered} \tag{14}$$

where $u$ and $v$ denote the velocity components in different directions, $\rho$ represents the density of the fluid, $v$ represents the kinematic viscosity coefficient of the fluid, and $Re$ is the Reynolds number, which reflects the relative magnitude of inertial forces and viscous forces in fluid flow.

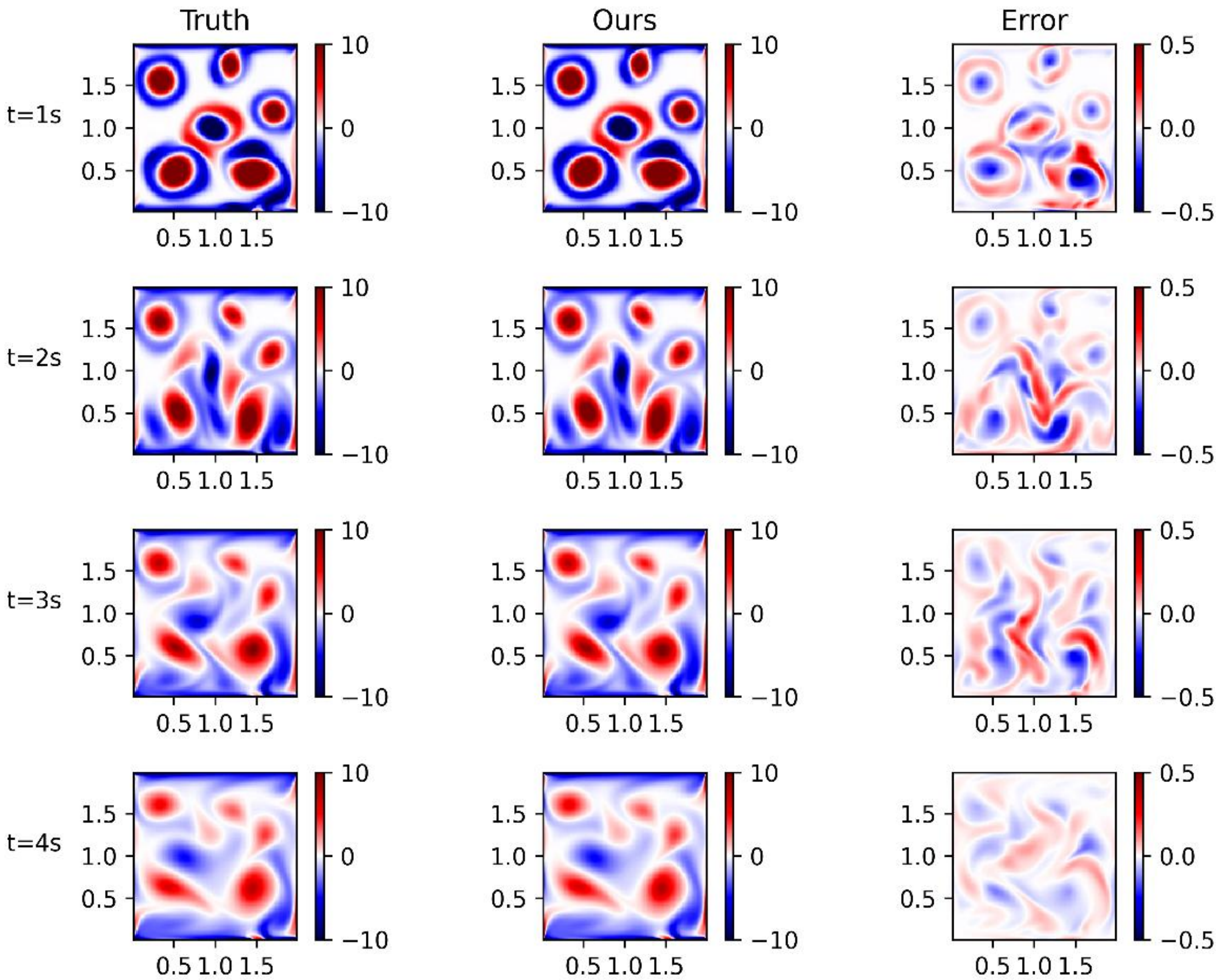

**Figure 19**. Results of Navier-Stokes Equation when $Re = 1000$.

In incompressible fluid dynamics, the Reynolds number plays a central role in determining the qualitative behavior of the flow. As Re increases, viscous dissipation becomes weaker relative to convective transport, leading to sharper velocity gradients, more pronounced vortical structures, and stronger multi-scale interactions. Consequently, flows at higher Reynolds numbers are generally more challenging to model and predict, particularly for learning-based methods. To further characterize the rotational structures of the flow, we consider the vorticity field. Vorticity is a fundamental quantity in fluid mechanics that describes the local rotational motion of fluid elements and is defined as the curl of the velocity field. In two-dimensional settings, the vorticity reduces to a scalar quantity given by $\omega = \frac{\partial v}{\partial x} - \frac{\partial u}{\partial y}$. Here we consider the situation of velocity when $Re = 500, 1000$, 2000. In particular, we present the ground truth and our predicted solutions of the vorticity field $\omega$ of the two-dimensional Navier–Stokes equation at the specified $Re = 1000$, and the errors between them at four time points: $t = 1\text{s}$, $t = 2\text{s}$, $t = 3\text{s}$ and $t = 4\text{s}$ in **Figure 19**.

Lastly, we compare the proposed method with several existing classical approaches on three representative partial differential equations. The quantitative results are summarized in Table 1, where the reported values correspond to the average relative mean squared errors (MSE) with respect to the ground-truth solutions, computed over five independent runs. In Table 1, the labels S, M, and L denote *small*, *medium*, and *large* parameter regimes. Specifically, for the Allen–Cahn equation, the three regimes are defined by increasing values of the interface parameter $\varphi = 0.5,\ 1,\ 2$, where smaller $\varphi$ leads to thinner interfaces and stronger stiffness. For the Klein–Gordon equation, the regimes correspond to increasing values of the mass parameter $m$ = 1, 3, 5, which induce higher-frequency temporal oscillations. For the Navier–Stokes equation, the regimes are defined by increasing Reynolds numbers, with $\text{Re} = 500$, $1000$, and $2000$ for the S, M, and L cases, respectively, corresponding to progressively more convection-dominated and multi-scale flow conditions. It should be noted that we were unable to find publicly available implementations for FI-PINNs and AAS-PINNs. Therefore, we re-implemented these methods based on the algorithmic descriptions provided in the respective papers in order to conduct a fair comparison.

**Table 1.** Average relative MSE over five experiments of different networks facing three examples.

| Method | *AC-S* | *AC-M* | *AC-L* | *KG-S* | *KG-M* | *KG-L* | *NS-S* | *NS-M* | *NS-L* |
|---|---|---|---|---|---|---|---|---|---|
| Vanilla PINNs | 5.62e-01 | 3.82e-01 | 3.05e-01 | 1.52e-02 | 4.53e-02 | 1.06e-01 | 1.11e-02 | 2.72e-02 | 6.58e-02 |
| Time marching | 2.18e-02 | 1.45e-02 | 1.33e-02 | 6.61e-03 | 1.73e-02 | 4.02e-02 | 8.86e-02 | 1.45e-01 | 3.09e-01 |
| PINNs-RAR | 9.36e-02 | 5.71e-02 | 4.26e-02 | 4.21e-04 | 9.42e-04 | 1.58e-03 | 4.87e-03 | 1.16e-02 | 2.74e-02 |
| FI-PINNs | 1.69e-02 | 7.16e-03 | 5.44e-03 | 1.94e-03 | 5.39e-03 | 1.14e-02 | 2.63e-03 | 6.75e-03 | 1.79e-02 |
| AAS-PINNs | 6.41e-04 | 3.90e-04 | 2.98e-04 | 7.96e-04 | 1.58e-03 | 3.47e-03 | 1.65e-03 | 2.84e-03 | 3.55e-03 |
| Ours | 2.08e-04 | 1.36e-04 | 1.07e-04 | 3.88e-04 | 8.09e-04 | 1.71e-03 | 3.36e-04 | 7.14e-04 | 1.83e-03 |

To further analyze the contributions of individual components within the proposed network, we conducted ablation experiments on the Allen-Cahn Equation, and the specific results are provided in Table 2. During the experiments, we systematically removed key components such as the penalty and GAN modules and evaluated model performance under identical training conditions.

**Table 2**. Ablation study on the proposed framework (relative MSE).

| Method | *Allen–Cahn* | *Klein–Gordon* | *Navier–Stokes* |
|---|---|---|---|
| Original PINNs | 3.82e-01 | 4.53e-02 | 2.72e-02 |
| Time marching PINNs | 1.45e-02 | 1.73e-02 | 1.45e-01 |
| Phy-Transformer w/o penalty | 4.73e-03 | 2.49e-03 | 4.80e-03 |
| PhyTF-GAN w/o penalty | 1.19e-03 | 1.05e-03 | 1.41e-03 |
| Phy-Transformer | 2.98e-04 | 1.26e-03 | 2.77e-03 |
| PhyTF-GAN | 1.36e-04 | 8.09e-04 | 7.14e-04 |

The baseline methods "Original PINNs" and "Time marching PINNs" show high MSEs ($3.82\times10^{-1}$ and $1.45\times10^{-2}$), highlighting limitations in modeling sharp phase-transition interfaces. Ablation results reveal critical roles for penalty and GAN: removing penalty increases MSE to $4.73\times10^{-3}$, proving its essential role in enforcing temporal causality—vital for physical consistency in time-dependent dynamics. While on this basis, GAN could help to figure out the problematic domains, improving the performance to a much higher state.

**4 Conclusion**

This study presents a two-stage residual-guided training strategy that synergizes Physics-Informed Transformers with GANs to overcome key limitations of conventional PINNs. The proposed framework addresses two critical challenges: (1) the oversight of high-residual regions due to global loss averaging and (2) violations of temporal causality in time-dependent PDEs. By embedding a decoder-only Transformer with causal masking and integrating residual-aware adaptive sampling via GANs, our approach ensures physically consistent solutions while dynamically focusing on under-optimized regions.

Numerical experiments validate the method's superiority, achieving improved accuracy and robustness across multiple benchmark equations. For instance, on the Allen-Cahn Equation, the proposed framework attains a relative mean squared error of $1.36\times10^{-4}$, significantly outperforming baseline PINN-based approaches. Ablation studies further confirm the complementary roles of the two key components: the GAN-based sampling module enhances the identification of localized high-residual regions, while the causal penalty term effectively enforces temporal consistency and stabilizes long-range dependency learning. Despite these advantages, the proposed framework also has limitations. The increased architectural complexity introduced by the Transformer and GAN components leads to higher computational cost compared with standard PINN formulations. In addition, as GAN training is known to be sensitive to hyperparameter choices, adapting the framework to new PDE systems may require careful parameter calibration, which can limit out-of-the-box generalization. Future work will focus on extending the proposed framework to more complex multi-physics scenarios involving coupled PDE systems, where interactions among multiple physical fields pose additional challenges for maintaining global consistency. We also plan to explore more advanced adaptive sampling mechanisms, including reinforcement learning–based strategies, to further improve the identification and refinement of highly localized residual structures in high-dimensional spatiotemporal domains.

**Data Availability**

All the data and code involved in this paper, as well as supplementary materials, will be uploaded to https://github.com/macroni0321/PhyTF-GAN once this paper is accepted.

**Declaration of Competing Interests**

The authors declare that they have no known competing financial interests or personal relationships that could have appeared to influence the work reported in this paper.

**Acknowledgements**

The authors would like to thank the Mathematics Research Branch Institute of Beijing Association of Higher Education & Beijing Interdisciplinary Science Society, and also to the Beijing Students' Innovation and Entrepreneurship Training program.